\pdfoutput=1
\documentclass[11pt]{article}

\usepackage[table]{xcolor}

\usepackage[final]{acl}

\usepackage[T1]{fontenc}
\usepackage[utf8]{inputenc}
\usepackage{textcomp}
\usepackage{times}
\usepackage{inconsolata}
\usepackage{latexsym}

\usepackage{microtype}
\usepackage{adjustbox}
\usepackage{multicol}
\usepackage{booktabs}
\usepackage{multirow}
\usepackage{enumitem}
\usepackage{pdflscape}
\usepackage{tcolorbox}
\tcbuselibrary{listings, skins, breakable}

\usepackage{amsmath}
\usepackage{amsfonts}
\usepackage{amssymb}
\usepackage{bbm}

\usepackage{graphicx}

\usepackage{fontawesome5}

\newcommand{\emoji}[1]{\includegraphics[height=1em]{#1}}

\newtcolorbox{takeawaybox}{
  enhanced,
  breakable,
  colback=green!5!white,
  colframe=green!20!white,
  coltitle=black,
  fonttitle=\bfseries,
  title={\faLeaf\hspace{0.5em}Takeaway},
  boxrule=0.4pt,
  arc=3pt,
  left=8pt,
  right=8pt,
  top=6pt,
  bottom=6pt,
  before skip=10pt,
  after skip=10pt
}

\title{Will It Still Be True Tomorrow? \\
Multilingual Evergreen Question Classification to Improve Trustworthy QA}

\author{
 \textbf{Sergey Pletenev\textsuperscript{*,2,1}},
 \textbf{Maria Marina\textsuperscript{*, 2,1}},
 \textbf{Nikolay Ivanov\textsuperscript{1}},
 \textbf{Daria Galimzianova\textsuperscript{4}},
 \\
 \textbf{Nikita Krayko\textsuperscript{4}},
  \textbf{Mikhail Salnikov\textsuperscript{2,1}},
 \textbf{Vasily Konovalov\textsuperscript{2,5}},
\\
 \textbf{Alexander Panchenko\textsuperscript{1,2},
 \textbf{Viktor Moskvoretskii\textsuperscript{1,3}}}
\\
 \textsuperscript{1}Skoltech,
 \textsuperscript{2}AIRI,
 \textsuperscript{3}HSE University,
 \textsuperscript{4}MTS AI,
 \textsuperscript{5}MIPT
\\
\href{mailto:Maria.Marina@skol.tech}{\{Maria.Marina}, 
\href{mailto:A.Panchenko@skol.tech}{A.Panchenko}, 
\href{mailto:v.moskvoretskii@skol.tech}{V.Moskvoretskii\}}@skol.tech
}

\begin{document}
\maketitle
\begin{abstract}

\footnotetext[1]{* Equal contribution.}

Large Language Models (LLMs) often hallucinate in question answering (QA) tasks. A key yet underexplored factor contributing to this is the temporality of questions -- whether they are \texttt{evergreen} (answers remain stable over time) or \texttt{mutable} (answers change). In this work, we introduce \textbf{EverGreenQA}, the first multilingual QA dataset with evergreen labels, supporting both evaluation and training.
Using \textbf{EverGreenQA}, we benchmark 12 modern LLMs to assess whether they encode question temporality explicitly (via verbalized judgments) or implicitly (via uncertainty signals). We also train \textbf{EG-E5}, a lightweight multilingual classifier that achieves SoTA performance on this task. Finally, we demonstrate the practical utility of evergreen classification across three applications: improving self-knowledge estimation, filtering QA datasets, and explaining GPT-4o’s retrieval behavior.


\end{abstract}


\section{Introduction}

Large language models (LLMs) often struggle with question answering (QA) due to hallucinated answers~\cite{Huang_2025}. To improve trustworthiness, recent research has focused on estimating LLMs' \textit{self-knowledge} -- their ability to recognize what they do and do not know~\cite{yin2023large,moskvoretskii2025adaptive} -- and on integrating up-to-date external information through Retrieval-Augmented Generation (RAG)~\cite{DBLP:conf/acl/SuTA0024, DBLP:conf/naacl/JeongBCHP24, DBLP:conf/acl/TrivediBKS23}.

\begin{table*}[ht!]
\centering
\footnotesize
\setlength{\tabcolsep}{2pt}
\resizebox{0.95\textwidth}{!}{

\begin{tabular}{lccccc}
\toprule
& \textbf{EverGreenQA} & \textbf{TimeQA} & \textbf{MuLan} & \textbf{FreshQA} & \textbf{TAQA} \\
& (our work) & \cite{Chen2021ADF} & ~\cite{DBLP:conf/naacl/FierroGBKS24} & \cite{DBLP:conf/acl/VuI0CWWTSZLL24} & \cite{DBLP:conf/acl/ZhaoBWHS24}  \\
\midrule
Both EG and mutable questions& \hspace{2.4ex} \emoji{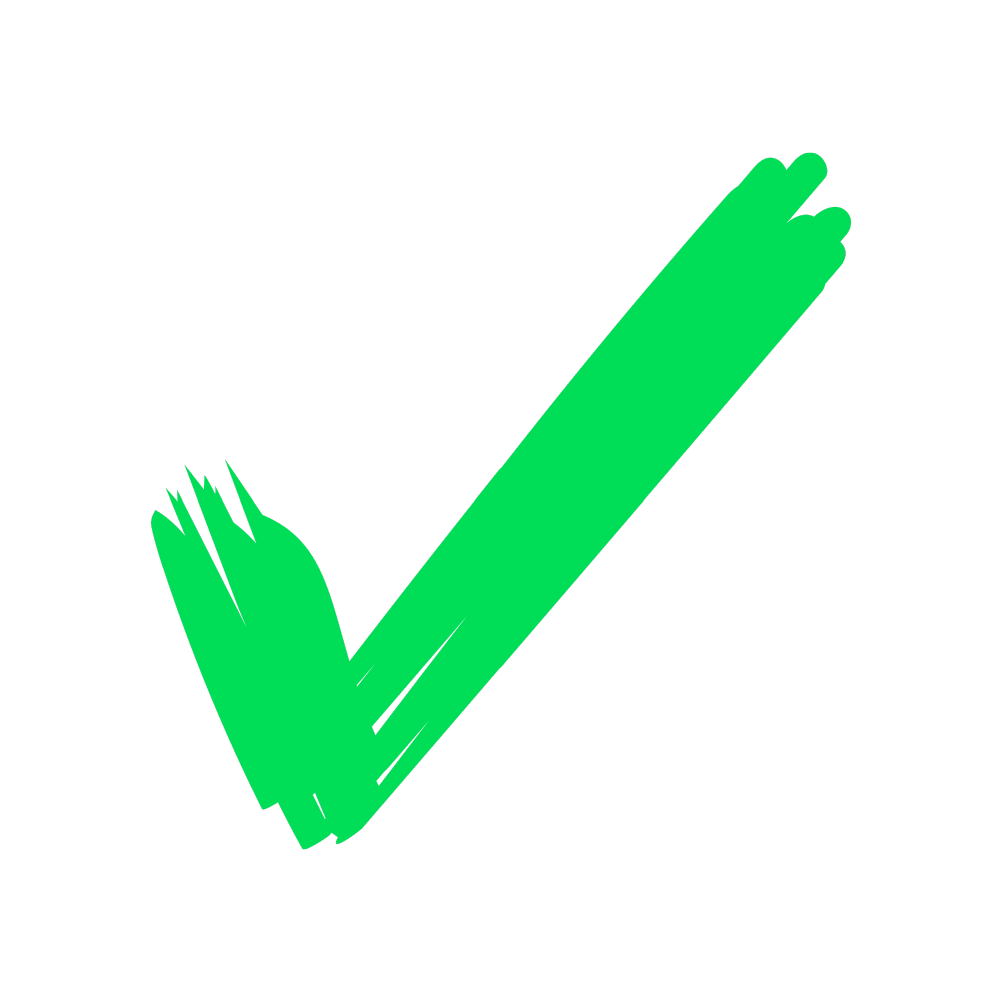} & \hspace{2.4ex} \emoji{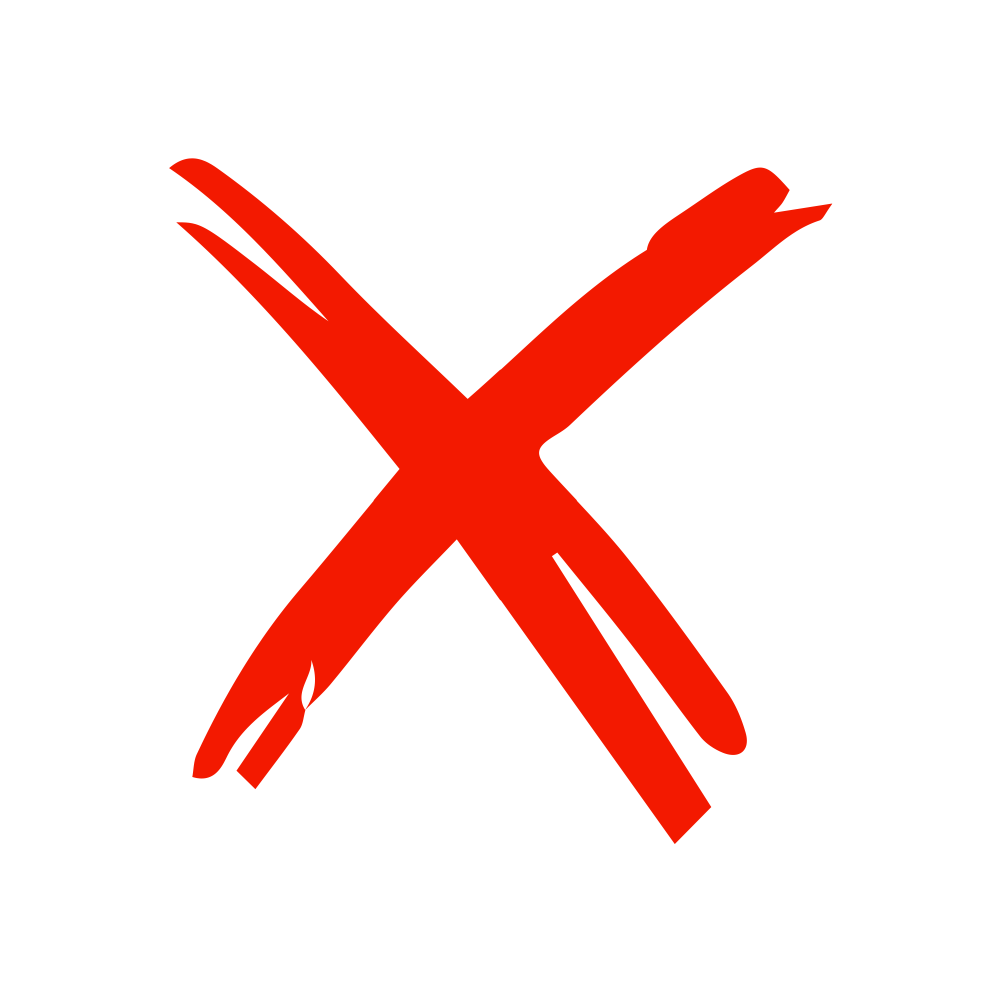}& \hspace{2.2ex} \emoji{Images/green_ok.pdf} & \hspace{2.2ex} \emoji{Images/green_ok.pdf} & \hspace{2.2ex} \emoji{Images/red_cross.pdf}\\

Train-Test split & \hspace{2.5ex} \emoji{Images/green_ok.pdf} & \hspace{2.4ex} 
 \emoji{Images/green_ok.pdf} & \hspace{2.2ex} \emoji{Images/red_cross.pdf}& \hspace{2.2ex} \emoji{Images/red_cross.pdf} & \hspace{2.4ex} 
 \emoji{Images/green_ok.pdf} \\

Human-Evaluated & \hspace{2.5ex} \emoji{Images/green_ok.pdf} & \hspace{2.5ex} \emoji{Images/green_ok.pdf} & \hspace{2.2ex} \emoji{Images/red_cross.pdf}& \hspace{2.5ex} \emoji{Images/green_ok.pdf}  & \hspace{2.2ex} \emoji{Images/red_cross.pdf} \\

Multilinguality & \hspace{2.5ex} \emoji{Images/green_ok.pdf} & \hspace{2.5ex} \emoji{Images/red_cross.pdf} & \hspace{2.2ex} \emoji{Images/red_cross.pdf}& \hspace{2.5ex} \emoji{Images/red_cross.pdf}  & \hspace{2.2ex} \emoji{Images/red_cross.pdf} \\

\midrule
Overall size & 4757 & \textasciitilde
 40k & \textasciitilde
246k & 600 &  \textasciitilde
20k\\
\bottomrule
\end{tabular}
}
\caption{Comparison of EverGreenQA to other time-sensitive datasets. }
\label{tab:datasets comparison}
\end{table*}

\begin{figure}[t!]
    \centering
    \includegraphics[width=\linewidth]{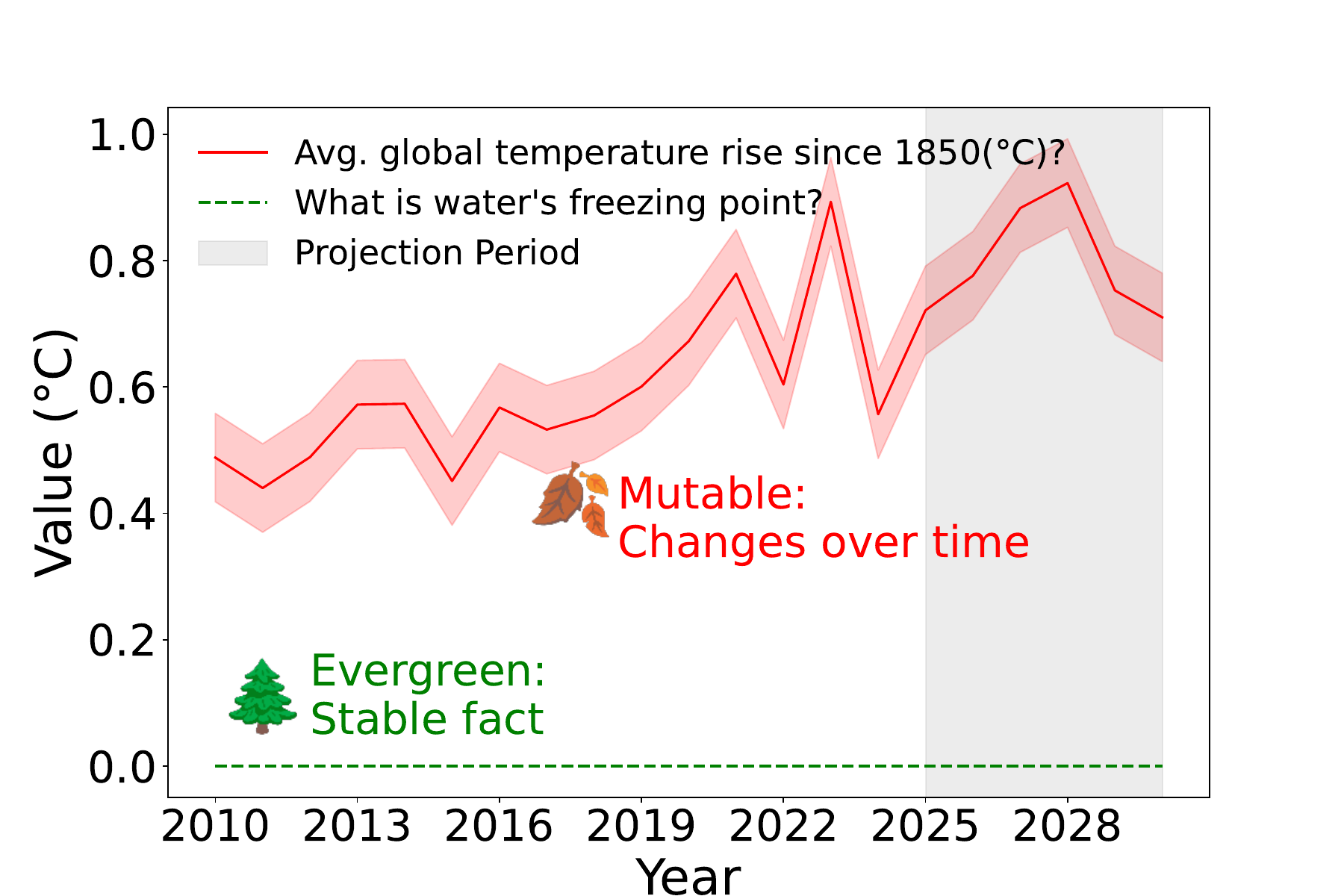}
    \caption{Some questions have answers that stay the same (\textbf{evergreen}), like facts of nature. Others have answers that change over time and will change in the future (\textbf{mutable}), like global trends or statistics.
    }
    \label{fig:evergreen}
    \vspace{-0.2cm}
\end{figure}

A particularly important but underexplored factor affecting question difficulty is whether a question is \texttt{evergreen} or \texttt{mutable}~\cite{wei2024measuringshortformfactualitylarge} -- that is, whether its correct answer remains stable over time, depicted with illustrative Figure~\ref{fig:evergreen}. Mutable questions are especially challenging because they often require access to up-to-date information, which may be missing from a model’s fixed, parametric knowledge.

Despite its practical importance, evergreen-ness remains an underexplored factor in evaluating and improving LLM behavior. Most existing studies are limited to small-scale, English-only datasets and focus primarily on QA accuracy, rarely examining its broader implications~\cite{DBLP:conf/acl/VuI0CWWTSZLL24, cheng2024unifiedactiveretrievalretrieval}. As a result, the role of question evergreen-ness in shaping LLM reliability and interpretability remains largely unexamined.

To address this gap, we conduct a comprehensive study of question evergreen-ness and its practical applications.{ We introduce \textbf{EverGreenQA} - the first multilingual human-curated evergreen-aware QA dataset}, which includes a train–test split suitable for model training. Using \textbf{EverGreenQA}, we evaluate 12 modern LLMs to determine whether they encode temporal knowledge explicitly (through direct prompting) or implicitly (via uncertainty-based signals). Further, we develop \textbf{EG-E5} -- a lightweight SoTA classifier trained to identify evergreen questions.

We demonstrate the usefulness of \textbf{EG-E5} in several downstream tasks: (1) improving self-knowledge estimation, (2) curating QA datasets to support fairer evaluation, and (3) effectively explaining GPT-4o’s black-box retrieval behavior.

Our contributions and findings are as follows: 

\begin{enumerate}[itemsep=0.5pt, topsep=0.5pt]
    \item We construct \textbf{EverGreenQA} -- the first multilingual dataset for question evergreen-ness classification, covering 7 languages with 4,757 examples in total. 
    
    \item We conduct the first comprehensive evaluation of question evergreen knowledge in LLMs, assessing 12 models using both explicit signals (via prompting) and implicit signals (via uncertainty estimation).
    
    \item We develop \textbf{EG-E5} -- a multilingual lightweight classifier for identifying evergreen questions, which serves as SoTA approach for question evergreen-ness classification while remaining suitable for low-compute settings.
    
    \item We demonstrate the utility of \textbf{EG-E5} across three applications: (1) improving self-knowledge estimation, (2) curating QA datasets for fairer evaluation, and (3) effectively explaining GPT-4o's retrieval behavior.
\end{enumerate}

We release the model and data for further usage.\footnote{\url{https://huggingface.co/collections/s-nlp/evergreen-683465909575cb89d6b904fe}}

\section{Related Work}
Reasoning about time remains a fundamental challenge in question answering (QA) tasks, as temporal dynamics often complicate both the interpretation of questions and the retrieval of accurate answers. Working with time in QA tasks has been improved thanks to datasets like 
\textit{TimeQA} \cite{Chen2021ADF}, which has 20,000 question-answer pairs that need thinking about time. While helpful, it only addressed simple reasoning. \textit{SituatedQA} \cite{DBLP:conf/emnlp/ZhangC21} showed the importance of context by situating questions in time and place. \textit{StreamingQA} highlighted the need for temporal adaptation, revealing LLMs' difficulty tracking changing facts~\cite{pmlr-v162-liska22a}. \textit{TemporalAlignmentQA~(TAQA)}~\cite{DBLP:conf/acl/ZhaoBWHS24} further enhances possibility for temporal alignment by providing 20K time-sensitive questions
and their answers for each year from 2000 to 2023. \textit{MuLan}~\cite{DBLP:conf/naacl/FierroGBKS24} differentiated questions by change rate and fact type, respectively. Most recently, \textit{FreshQA}~\cite{DBLP:conf/acl/VuI0CWWTSZLL24} introduced a benchmark focused on freshness-sensitive information, further illustrating LLMs' limitations in handling temporally dynamic knowledge. These studies indicate a need for specialized temporal reasoning~\cite{DBLP:conf/naacl/FierroGBKS24}. The comparison of datasets is presented in Table~\ref{tab:datasets comparison}.

Retrieval-Augmented Generation (RAG), such as DRAGIN~\cite{DBLP:conf/acl/SuTA0024}, IRCoT~\cite{DBLP:conf/acl/TrivediBKS23} or Rowen~\cite{DBLP:journals/corr/abs-2402-10612-rowen} was used to solve the problem of time sensitive QA, addressed this through dynamic retrieval decisions, but showed limited results. 

Dynamic retrieval decisions required self-knowledge estimation. In other words, before QA systems can be trusted, they need to know what they don't know. Often, LLMs struggle to identify questions they can't answer~\cite{yin2023large}, but using self-knowledge can reduce mistakes in tasks that need a lot of knowledge~\cite{DBLP:conf/emnlp/WangLSL23, moskvoretskii2025adaptive}.

While retrieval-based methods address temporal knowledge gaps externally, another direction is to update the internal knowledge of LLMs. Updating internal knowledge in LLMs is computationally expensive, as retraining or editing models often requires substantial resources and cannot be performed daily or hourly in practice. Techniques like \textit{LLM Surgery}~\cite{veldanda2024llm} and parameter-efficient fine-tuning~\cite{ge-etal-2024-time, pletenev-etal-2025-much} have attempted to make such updates more practical, but still face issues with large-scale changes or factual hallucinations.

\begin{table*}[ht!]
\centering
\footnotesize
\resizebox{\textwidth}{!}{
\begin{tabular}{lcccccccc}
\toprule
\textbf{Model}                      & \textbf{Russian} & \textbf{English} & \textbf{French} & \textbf{German} & \textbf{Hebrew} & \textbf{Arabic} & \textbf{Chinese} & \textbf{AVG}   \\

\midrule

LLaMA 3.1-8B-it              & 0.677  & 0.699  & 0.686  & 0.677  & 0.667   & 0.659   & 0.652    & 0.674     \\
LLaMA 3.1-70B-it             & 0.889   & 0.879  & 0.895 & 0.87   & 0.874  & 0.829    & 0.873    & 0.875 \\
Qwen 2.5 7B-it                & 0.782  & 0.789    & 0.786   & 0.794   & 0.692   & 0.711   & 0.774    & 0.761 \\
Qwen 2.5 32B-it                & 0.882    & 0.885   & 0.875   & 0.883   & 0.862   & 0.862   & 0.872    & 0.874 \\
Qwen 2.5 72B-it                & 0.806    & 0.815    & 0.802   & 0.805   & 0.781   & 0.758   & 0.768  & 0.791 \\
Phi-3 medium 4k-it  & 0.556 & 0.577  & 0.498 & 0.473 & 0.499  & 0.498  & 0.420  & 0.503 \\
Phi-3 medium 128k-it    & 0.415    & 0.489    & 0.342   & 0.335   & 0.385   & 0.304   & 0.289  & 0.366 \\
Gemma 2-9B-it  & 0.755 & 0.728 & 0.694 & 0.723 & 0.740 & 0.711 & 0.746 & 0.728 \\
Gemma 2-27B-it & 0.830 & 0.878 & 0.836 & 0.827 & 0.838 & 0.831 & 0.826 & 0.838 \\
Mistral 7B-it-v0.3  & 0.736 & 0.722 & 0.726 & 0.729 & 0.670 & 0.666 & 0.731 & 0.711 \\
Mistral Small-24B-it-2501  & 0.827 & 0.739 & 0.768 & 0.789 & 0.847 & 0.834 & 0.839 & 0.806 \\
GPT-4.1                    & 0.806  & 0.794 & 0.816 & 0.813  & 0.803 & 0.811  & 0.809   & 0.807 \\
\midrule
UAR~\cite{cheng2024unifiedactiveretrievalretrieval}          & 0.550    & 0.500    & 0.510   & 0.600   & 0.670   & 0.710   & 0.710    & 0.490     \\

MULAN~\cite{DBLP:conf/naacl/FierroGBKS24}        & 0.340 & 0.345 & 0.442  & 0.379  & 0.322  & 0.220  & 0.279    & 0.340     \\
\textbf{EG-E5 (our)} & \textbf{0.910}   & \textbf{0.913}   & \textbf{0.909}  & \textbf{0.910}  & \textbf{0.904}  & \textbf{0.900}  & \textbf{0.897}   & \textbf{0.906}    \\
\bottomrule
\end{tabular}
}
\caption{Comparison of verbalized LLM predictions and our trained classifier on the test part of evergreen classification task. Reported scores are weighted F1. A random baseline achieves 0.637. LLMs were prompted with 10-shot examples. The best scores are shown in \textbf{bold}.}
\label{tab:verbal}
\end{table*}

\section{EverGreenQA \& EG-E5}

\textbf{Dataset Collection.} We construct a QA dataset consisting of real user queries sourced from an AI chat assistant, each labeled as either \texttt{evergreen} or \texttt{mutable}, along with corresponding golden answers. All questions are factual in nature and were manually validated over multiple iterations of internal alpha testing to ensure diversity and reduce topic bias. The labels and golden answers were assigned by a team of trained linguists, who manually wrote the answers from scratch based on retrieved information. Due to the fact that in the initial dataset most of the questions were \texttt{mutable} and to avoid bias in the training data, we also generated 1,449 synthetic data for the \texttt{evergreen} class only. This additional dataset was similarly validated by linguists. The final dataset contains 4,757 questions, with 3,487 used for training and 1,270 reserved for testing, details of dataset collection and labeling are presented in Appendix~\ref{appendix:data_collection_details}.

\textbf{Dataset Translation.} 
We perform translations from Russian to English and from English to the target languages using GPT-4.1, following prior work that demonstrated its strong performance across a wide range of languages, including accurate handling of cultural nuances~\cite{vayani2024all}. The full translation prompt is provided in Appendix~\ref{appendix:translation_prompt}.

\textbf{Dataset Validation.} 
To assess translation quality, we recruited human evaluators for each target language, all of whom are either native speakers or possess advanced proficiency (B2–C1 level). We randomly sampled 100 questions from the test set (50 mutable, 50 evergreen) for evaluation. No errors were found in the translations for English, Hebrew, German, or Arabic, while Chinese exhibited only two minor inaccuracies. Validation assessor instruction is provided in Appendix~\ref{appendix:validation}.

\textbf{EG-E5 Training.}
For training and testing, we used our multilingual dataset. For validation, we employed the dev and test splits from FreshQA~\cite{DBLP:conf/acl/VuI0CWWTSZLL24}, merging the \texttt{fast-changing} and \texttt{slow-changing} classes into \texttt{mutable} label. To align with our multilingual setting, the FreshQA data was translated into all target languages.

We experimented with multilingual versions of BERT~\cite{devlin2019bertpretrainingdeepbidirectional}, DeBERTaV3~\cite{he2023debertav3improvingdebertausing}, and E5~\cite{wang2024multilinguale5textembeddings} as encoders. The best performance was achieved using the E5-Large model, which we refer to as our classifier \textbf{EverGreen-E5 (EG-E5)}. Hyperparameter details and ablation results are provided in Appendix~\ref{appendix:model_evergreen}.

\section{Are LLMs Aware of Evergreenness?}

In this section, we evaluate whether modern LLMs can reliably identify whether a given question is evergreen. We test 12 LLMs spanning diverse architectures, with full details provided in Appendix~\ref{appendix:model_evergreen}. 

\subsection{Verbalized Evergreen Awareness}

To assess whether LLMs are capable of explicitly recognizing evergreen questions, we prompt each model to provide a binary Yes/No answer.

We additionally include two specifically trained methods: \textbf{UAR}~\cite{cheng2024unifiedactiveretrievalretrieval}: a previously proposed LLaMA2-13b fine-tuned to classify evergreen questions, and \textbf{MULAN}~\cite{DBLP:conf/naacl/FierroGBKS24} classification based on mutable and evergreen samples from Wikidata.

\textbf{Results.} Table~\ref{tab:verbal} shows that our proposed classifier, \textbf{EG-E5}, achieves the highest performance across all languages, significantly outperforming both general-purpose and specifically trained LLMs. Among the LLMs, {LLaMA 3.1 70B} and {Qwen 2.5 32B} are the strongest, with GPT-4.1 lagging a bit behind.

We observe some variations in language performance, but no clear performance gap, even for non-Latin languages (e.g., Arabic, Chinese, Russian).

Baseline methods {UAR} and {MULAN} perform substantially worse than both LLMs and {EG-E5}, likely due to their oversimplified assumptions regarding the evergreen nature of QA datasets.

\begin{takeawaybox}
EG-E5 outperforms few-shot LLMs and prior methods, whose weaker results stem from unrealistic assumptions in their training data.
\end{takeawaybox}

\subsection{Internal Evergreen Awareness} \label{sec:internal}
We next assess whether LLMs implicitly encode information about question evergreen-ness through their uncertainty estimates using a balanced subset of sampled 400 questions from our test set -- 200 labeled as \texttt{evergreen} and 200 as \texttt{mutable}.

We select two widely adopted uncertainty measures that show strong performance~\cite{10.1162/tacl_a_00737,moskvoretskii2025adaptive}.

\textbf{Perplexity} -- the inverse probability of the predicted sequence, normalized by its length. For a sequence of tokens \( x_1, \dots, x_T \), it is defined as:
    \[
    \text{PPL} = \exp\left(-\frac{1}{T} \sum_{t=1}^{T} \log p(x_t \mid x_{<t})\right)
    \]

\textbf{Mean Token Entropy} -- the average entropy of the model’s predicted token distribution at each position:
    \[
    \text{Entropy} = -\frac{1}{T} \sum_{t=1}^{T} \sum_{w \in V} p_t(w) \log p_t(w)
    \]
    where \( p_t(w) \) is the predicted probability of token \( w \) at position \( t \), and \( V \) is the vocabulary.

\begin{table}[t!]
\centering
\resizebox{0.48\textwidth}{!}{
\begin{tabular}{lcc}
\toprule
{\textbf{Model}} & {\textbf{Perplexity}} & {\textbf{Mean Token Entropy}} \\
\midrule
Gemma 2-9B-it        & 0.23                & 0.27               \\
Gemma 2-27B-it       & 0.26               & 0.29                \\
LLaMA 3.1-8B-it       & 0.33 & 0.33 \\
LLaMA 3.1-70B-it       & 0.20  & 0.21   \\
Mistral 7B-it-v0.3         & 0.33  & 0.35                \\
Mistral Small-24B-it-2501          & 0.34 & 0.32  \\
Phi-3 medium 4k-it            & 0.23  & 0.17   \\
Phi-3 medium 128k-it           & 0.27 & 0.32 \\
Qwen 2.5 7B-it          & 0.25  & 0.25  \\
Qwen 2.5 32B-it         & 0.33   & 0.34 \\
Qwen 2.5 72B-it        & 0.29  & 0.31 \\
\bottomrule
\end{tabular}
}
\caption{Correlation of golden EG with UC. All results are significant (p-value < 0.05).}
\label{tab:corr_EG_UC}
\end{table}

\textbf{Results.} 
Table~\ref{tab:corr_EG_UC} shows that most models exhibit only mild correlations between uncertainty and evergreen-ness, with Mistral 7B and Qwen 2.5 32B achieving the strongest signals.

We also observe a weak trend suggesting that larger models correlate more strongly with evergreen-ness, possibly indicating a greater internal reliance on temporal cues. Neither perplexity nor entropy consistently outperforms the other. Overall, uncertainty signals capture some temporal information, but are noticeably weaker than explicit verbalized judgments. Additional analysis is provided in Appendix~\ref{appendix:r2_ue}.

\begin{takeawaybox}
Uncertainty metrics encode weak and inconsistent signals of evergreen-ness, with slightly stronger trends in larger models.
\end{takeawaybox}

\begin{table*}[ht!]
\centering
\setlength{\tabcolsep}{2pt}
\resizebox{\textwidth}{!}{
\begin{tabular}{l|ccc|ccc|ccc|ccc|ccc|ccc}
\toprule
& \multicolumn{3}{c|}{\textbf{NQ}} & \multicolumn{3}{c|}{\textbf{SQuAD}} & \multicolumn{3}{c|}{\textbf{TriviaQA}} & \multicolumn{3}{c|}{\textbf{2WikiMultihopQA}} & \multicolumn{3}{c|}{\textbf{HotpotQA}} & \multicolumn{3}{c}{\textbf{MuSiQue}} \\

\textbf{Method} & AUROC & AUPRC & PRR & AUROC & AUPRC & PRR & AUROC & AUPRC & PRR & AUROC & AUPRC & PRR & AUROC & AUPRC & PRR & AUROC & AUPRC & PRR \\
\midrule
\multicolumn{19}{c}{\textit{Uncertainty Estimation}} \\
\midrule
EigValLaplacian       & 0.56 & 0.46 & 0.56 & 0.48 & 0.19 & 0.77 & 0.70 & 0.52 & 0.58 & 0.61 & 0.26 & 0.71 & 0.64 & 0.22 & 0.75 & 0.57 & 0.09 & 0.86 \\
LexicalSimilarity     & \textbf{0.61} & 0.38 & 0.59 & 0.64 & 0.13 & 0.83 & 0.65 & 0.54 & 0.58 & 0.55 & 0.29 & 0.67 & 0.68 & 0.21 & 0.77 & 0.58 & 0.09 & 0.85 \\
MaxTokenEnt.          & \textbf{0.61} & 0.37 & \textbf{0.60} & 0.58 & 0.18 & 0.80 & 0.70 & 0.51 & 0.62 & 0.59 & 0.27 & 0.69 & 0.67 & 0.21 & 0.75 & 0.64 & 0.09 & 0.84 \\
MeanTokenEnt.         & 0.59 & 0.42 & 0.57 & 0.56 & 0.19 & 0.80 & 0.71 & 0.50 & \textbf{0.63} & 0.61 & 0.28 & 0.70 & 0.62 & 0.23 & 0.73 & 0.63 & 0.09 & 0.81 \\
SAR                   & \textbf{0.61} & 0.39 & 0.59 & 0.67 & 0.12 & 0.84 & \textbf{0.72} & 0.51 & 0.60 & 0.60 & 0.27 & 0.69 & 0.69 & 0.21 & \textbf{0.78} & 0.64 & 0.08 & 0.83 \\
\midrule
\multicolumn{19}{c}{\textit{Uncertainty Estimation + Evergreen}} \\
\midrule
EigValLaplacian+EG    & 0.56 & 0.40 & 0.57 & 0.49 & 0.19 & 0.77 & 0.70 & 0.51 & 0.56 & 0.54 & \textbf{0.52} & 0.64 & 0.65 & 0.21 & 0.75 & 0.50 & \textbf{0.12} & 0.84 \\
LexicalSimilarity+EG  & 0.59 & 0.40 & 0.59 & 0.65 & 0.13 & 0.83 & 0.68 & 0.52 & \textbf{0.63} & 0.61 & 0.26 & 0.71 & 0.68 & 0.21 & 0.76 & 0.61 & 0.10 & 0.86 \\
MaxTokenEnt.+EG       & 0.56 & 0.42 & 0.59 & 0.68 & 0.12 & 0.85 & 0.71 & 0.51 & \textbf{0.63} & \textbf{0.63} & 0.25 & \textbf{0.72} & 0.67 & 0.21 & 0.75 & 0.55 & 0.11 & 0.84 \\
MeanTokenEnt.+EG      & 0.59 & 0.39 & 0.59 & \textbf{0.70} & 0.12 & \textbf{0.86} & \textbf{0.72} & 0.50 & 0.62 & 0.61 & 0.26 & 0.71 & 0.63 & 0.22 & 0.75 & 0.64 & 0.08 & 0.85 \\
SAR+EG                & 0.58 & 0.41 & 0.57 & \textbf{0.70} & 0.12 & 0.85 & 0.66 & 0.54 & 0.46 & 0.62 & 0.43 & 0.68 & \textbf{0.70} & 0.21 & \textbf{0.78} & \textbf{0.67} & 0.11 & \textbf{0.87} \\
\midrule
\multicolumn{19}{c}{\textit{Evergreen}} \\
\midrule
EG                    & 0.50 & \textbf{0.72} & 0.52 & 0.52 & \textbf{0.20} & 0.79 & 0.47 & \textbf{0.65} & 0.62 & 0.49 & 0.31 & 0.65 & 0.51 & \textbf{0.28} & 0.68 & 0.50 & 0.10 &\textbf{0.87} \\
\bottomrule
\end{tabular}
}
\caption{Self-knowledge identification performance. We report classification quality using AUROC and AUPRC, and calibration efficiency using PRR. EG stand for Evergreen probability. Higher values indicate better performance. The best scores for each metric are shown in \textbf{bold}.}
\label{tab:AR_acc}
\end{table*}

\section{Enhancing Self-Knowledge} \label{sec:self_knowledge}
In this section, we evaluate whether incorporating knowledge about question evergreen-ness improves the estimation of \textit{self-knowledge} -- a model’s ability to recognize the boundaries of its own knowledge and determine when it can or cannot answer a given question~\cite{moskvoretskii2025adaptive,yin2023large}. This capability is considered a key factor in improving the trustworthiness of LLMs.

\subsection{Task Formulation}

We frame self-knowledge estimation as a binary classification task, where the target label \( y \in \{0, 1\} \) reflects whether the model's answer to a given input \( x \) is factually correct. Each method under evaluation assigns a real-valued self-knowledge score \( f(x) \in \mathbb{R} \) to the input.

\subsection{Methods}

We evaluate this setup using LLaMA3.1-8B-Instruct with five widely adopted and high-performing uncertainty estimators, selected to represent different families of uncertainty quantification methods -- including logit-based and consistency-based approaches:

\textbf{Max Token Entropy}: Evaluates uncertainty by computing token-level entropies and taking the maximum value across the sequence as the final score~\cite{fomicheva2020unsupervised}.

\textbf{Mean Token Entropy}: Similar to the above, but aggregates across the sequence by averaging token-level entropy values~\cite{fomicheva2020unsupervised}.

\textbf{Lexical Similarity}: Estimates uncertainty by calculating the average lexical overlap among multiple model responses, serving as a proxy for output consistency~\cite{fomicheva2020unsupervised}.
        
\textbf{SAR}: Combines entropy with relevance weighting by amplifying the contribution of semantically important tokens, summing the adjusted entropy values over the sequence~\cite{duan2023shifting}.

\textbf{EigValLaplacian}: Constructs a similarity graph over sampled responses and computes the sum of eigenvalues of its Laplacian matrix to quantify response diversity~\cite{lin2023generating}.

For each method, we evaluate the effect of incorporating the predicted probability of a question being evergreen, obtained from our trained evergreen classifier.

To obtain the final self-knowledge classifier \( f(x) \), we train a standard machine learning model on the training set, using the uncertainty estimation metrics as input features. When applicable, we also include the predicted evergreen probability as an additional feature. The full training procedure is detailed in Appendix~\ref{appendix:training_classifier}.

\begin{table*}[ht!]
\centering
\setlength{\tabcolsep}{2pt}
\resizebox{\textwidth}{!}{
\begin{tabular}{lllll}
\toprule
\textbf{Dataset} & \textbf{\begin{tabular}[c]{@{}l@{}}Dataset \\ release\end{tabular}} & \multicolumn{1}{c}{\textbf{Non-EG question}}                                                                                                    & \textbf{\begin{tabular}[c]{@{}l@{}}Reference \\ answer\end{tabular}} & \textbf{Answer in 2025} \\ \midrule
\multicolumn{5}{c}{\textit{ \emoji{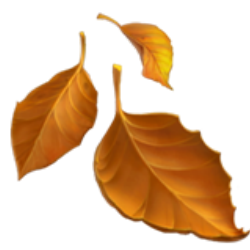}  
\emoji{Images/fallen_leaf.pdf}
\emoji{Images/fallen_leaf.pdf}
\emoji{Images/fallen_leaf.pdf}
\emoji{Images/fallen_leaf.pdf}
Expected changes (routine, scheduled)}}                                                                                                      \\ \midrule
NQ    & 2015   & what city is the next winter olympics in        & Beijing                   & Milan                   \\
MuSiQue          & 2022                                                                & Who is the mayor presiding now where Merrill Elam was born?                                                                                     & Lance Bottoms             & Andre Dickens           \\
SQuAD           & 2020                                                                & How many teams are in the Greek Super League?                                                                                                   & 18                        & 14                      \\
HotpotQA         & 2018                                                                & Yau Ma Tei North is a district of a city with how many citizens?                                                                                & 7.2 million               & 7.4 million             \\
MuSiQue          & 2022                                                                & \begin{tabular}[c]{@{}l@{}}According to QS World University Rankings, where does the college\\ that Ibrahim Shihata attended rank?\end{tabular} & 551-600                   & 350                     \\ \midrule
\multicolumn{5}{c}{\textit{ \emoji{Images/fallen_leaf.pdf}
\emoji{Images/fallen_leaf.pdf}
\emoji{Images/fallen_leaf.pdf}
Occasional changes (updates over time, but less regular)}}                                                                                                                                                                                                       \\ \midrule
HotpotQA         & 2018                                                                & Edoardo Soleri is playing on loan from which Italian football club?                                                                             & A.S. Roma                 & Spezia                  \\
2WikiMultihopQA            & 2020                                                                & Where does Karin Stoltenberg's husband work at?                                                                                                 & United Nations            & He has died             \\
2WikiMultihopQA            & 2020                                                                & Who is the spouse of the performer of song Les Rois Du Monde?                                                                                   & Joy Esther                & Emily Surde             \\
TriviaQA       & 2017                                                                & What is the name of the current Attorney General for England and Wales?                                                                         & Dominick Grieve           & Richard Hermer          \\
NQ               & 2015                                                                & who is the current minister for environment forest and climate change in india                                                                  & Dr. Harsh Vardhan         & Bhupender Yadav         \\ \midrule
\multicolumn{5}{c}{\textit{\emoji{Images/fallen_leaf.pdf} Less predictable changes (complex sociopolitical shifts)}}                                                                                                                                                                                                        \\ \midrule
SQuAD            & 2020                                                                & What is the largest economy in Africa?                                                                                                          & Nigeria                   & South Africa            \\
TriviaQA       & 2017                                                                & Who is fifth in line to the throne?                                                                                                             & Princess Beatrice         & Prince Harry            \\ \bottomrule
\end{tabular}
}
\caption{Examples of non-evergreen questions from popular QA datasets, showing discrepancies between original gold answers and updated answers in 2025. Questions are categorized by the nature of the change: expected, occasional, and less predictable. }
\label{tab:non_EG_changed}
\end{table*}

\subsection{Evaluation}

We evaluate performance using standard metrics widely adopted in recent literature on uncertainty estimation~\cite{fadeeva2024fact,10.1162/tacl_a_00737,vazhentsev2025token}.

\textbf{AUROC} measures how well the model distinguishes between correct and incorrect answers based on the self-knowledge score \( f(x) \). Higher values indicate stronger separability.

\textbf{AUPRC} quantifies the trade-off between precision and recall across different decision thresholds. It is particularly informative when dealing with imbalanced datasets.
    
\textbf{Prediction Rejection Ratio (PRR)} measures how well uncertainty scores align with answer quality. It simulates rejecting the most uncertain responses and tracks how average quality improves. Higher PRR indicates better calibration between uncertainty and actual answer correctness. We use In-Accuracy as main QA metric.

\subsection{Datasets} \label{sec:data}

We evaluate our methods on 6 QA datasets covering both single-hop and multi-hop reasoning. The single-hop datasets include SQuAD v1.1~\cite{DBLP:conf/emnlp/RajpurkarZLL16}, Natural Questions~\cite{DBLP:journals/tacl/KwiatkowskiPRCP19}, and TriviaQA~\cite{DBLP:conf/acl/JoshiCWZ17}, while the multi-hop datasets include MuSiQue~\cite{DBLP:journals/tacl/TrivediBKS22}, HotpotQA~\cite{DBLP:conf/emnlp/Yang0ZBCSM18}, and 2WikiMultiHopQA~\cite{DBLP:conf/coling/HoNSA20}. 

Following~\citet{DBLP:conf/acl/TrivediBKS23, DBLP:conf/naacl/JeongBCHP24}, we use a subset of 500 questions from the original splits of each dataset to ensure consistency and comparability.

\subsection{Results}

As shown in Table~\ref{tab:AR_acc}, evergreen probability is a strong signal for improving self-knowledge identification. In 16 out of 18 evaluations, the best results are achieved either by the evergreen feature alone or by combining it with an uncertainty estimation method. Moreover, it is able to improve calibration (PRR), that depends on QA accuracy, making it highly valuable for real-world applications.

Notably, the evergreen feature alone performs exceptionally well on AUPRC, achieving the top score on 4 datasets. This suggests that evergreen-ness is a powerful indicator of when a model possesses reliable knowledge.

However, we also observe a consistent pattern: evergreen scores high on AUPRC but relatively low on AUROC. This indicates that while the feature is highly effective at identifying when the model knows the answer, it is less reliable at recognizing when the model does not (weaker true negative discrimination). In other words, if a question is evergreen, the model is likely to answer it correctly -- but if a question is not evergreen, the outcome is harder to predict.

\begin{takeawaybox}
Evergreen probability consistently improves self-knowledge estimation and calibration, achieving top results in 16 out of 18 settings. 
\end{takeawaybox}

\section{Filtering QA with Evergreen}

In this section, we demonstrate that evergreen classification is valuable for filtering QA datasets, enabling fairer evaluation by excluding mutable questions. We use the same model setting as in Self-Knowledge Section~\ref{sec:self_knowledge}.

QA datasets should ideally consist only of evergreen questions, emphasized in SimpleQA~\cite{wei2024measuring}. To achieve this, SimpleQA relied on human annotators to assess evergreen-ness. In contrast, \textbf{EG-E5} enables automated dataset curation, eliminating the need for manual annotation and facilitating the scalable construction of large QA corpora. 

\begin{table*}[ht]
\centering
 \small

\begin{tabular}{lcccc|cc|c}
\toprule
\multirow{2}{*}{\textbf{Dataset}} & \multicolumn{2}{c}{\textbf{0-Shot}} & \multicolumn{2}{c|}{\textbf{RAG}} & {\textbf{$\Delta$ EG-Mut}} & \textbf{Mut RAG} & \multirow{2}{*}{\textbf{Mut \%}} \\
 & EG & Mut & EG & Mut &  \textbf{0-shot \%} &  \textbf{Gain \%} &  \\
\midrule
NQ & 0.399 & 0.344 & 0.660 & 0.635 & 16 & 10 & 18 \\
TriviaQA & 0.661 & 0.581 & 0.749 & 0.682 & 14 & 13 & 6 \\
SQuAD & 0.171 & 0.168 & 0.627 & 0.598 & 2 & -6 & 12 \\
HotpotQA & 0.367 & 0.282 & 0.746 & 0.727 & 30 & 14 & 10 \\
MuSiQue & 0.113 & 0.080 & 0.278 & 0.315 & 41 & 30 & 17 \\
\rowcolor{gray!30}

2wikiMultihopQA & 0.448 & 0.342 & 0.644 & 0.457 & 31 & -70 & 0.1 \\
\bottomrule
\end{tabular}

\caption{Performance comparison between evergreen (EG) and mutable (Mut) questions under 0-shot (no context) and RAG (with context) settings. 
We report absolute in-accuracies, the relative gap between evergreen and mutable questions ($\Delta$ EG–Mutable) under 0-shot, and the relative RAG gain on mutable questions. 
A higher mutable gain indicates RAG is more beneficial for time-sensitive queries. The last column shows the proportion of mutable questions in each dataset. 
\textcolor{gray!80!black}{Gray row indicates limited applicability due to extremely low mutable sample count.}}

\label{tab:ever_mut_rag_delta}
\end{table*}

\subsection{Popular QA Datasets Analysis}

Mutable questions pose a serious challenge for fair QA evaluation: outdated gold answers can make correct responses from modern LLMs appear wrong, especially when models are evaluated at different times.


\textbf{Examples.}
Table~\ref{tab:non_EG_changed} highlights such mutable examples across six datasets (Section~\ref{sec:data}), showing answers that, as of 2025, diverge from the original references. These include both simple and complex queries-even from recently released datasets like MuSiQue~\cite{DBLP:journals/tacl/TrivediBKS22}. The nature of change varies: some are predictable (e.g., Olympic host cities, population figures), some occasional (e.g., job titles or spouses), and others unexpected (e.g., monarchs, GDP rankings).

\textbf{Statistics.}
Table~\ref{tab:ever_mut_rag_delta} shows that mutable questions remain common, reaching 18\% in NQ and averaging 10\% across datasets. This challenges the widespread assumption that QA benchmarks are temporally stable, and raises concerns about evaluation fairness. To ensure reliability, mutable questions should be filtered out, or alternatively, live benchmarks like RealTimeQA~\cite{kasai2024realtimeqawhatsanswer} should be maintained-though they are costly to sustain.

\textbf{Incorrect Assumptions.} UAR~\cite{cheng2024unifiedactiveretrievalretrieval} has implicitly assumed dataset evergreen-ness and MULAN~\cite{DBLP:conf/naacl/FierroGBKS24} treat many questions as immutable, yet some relations (e.g., Wikidata’s P190, “sister cities”) can in fact change. This mismatch may help explain the limited real-world effectiveness of such methods when faced with temporal drift.

\begin{takeawaybox}
QA benchmarks include mutable questions, undermining fair evaluation. Filtering for evergreen questions is essential for reliable assessment.
\end{takeawaybox}

\subsection{Filtered QA Performance}

\textbf{Zero-Shot Performance.} As shown in Table~\ref{tab:ever_mut_rag_delta}, model accuracy is consistently higher on evergreen questions, with relative differences reaching up to 40\% on complex tasks. This aligns with expectations, as mutable questions often require up-to-date information beyond the model’s static knowledge.

\textbf{RAG Benefits.} We show that models generally benefit more from RAG with gold contexts when answering mutable questions, with relative gains reaching up to 30\%. However, this effect diminishes in datasets with few mutable examples.



\begin{table}[t!]
\centering
\small
\begin{tabular}{lcc}
\toprule
{Model} &  {ChatGPT} \\
\midrule
Gemma 2-9B-it                     & 0.26               \\
Gemma 2-27B-it            & 0.30                \\
LLaMA 3.1-8B-it      & 0.29 \\
LLaMA 3.1-70B-it      & 0.25   \\
Mistral 7B-it-v0.3       & 0.34                \\
Mistral Small-24B-it-2501     & 0.33  \\
Phi-3 medium 4k-it            & 0.20   \\
Phi-3 medium 128k-it            & 0.29 \\
Qwen 2.5 7B-it           & 0.28  \\
Qwen 2.5 32B-it         & 0.36 \\
Qwen 2.5 72B-it       & 0.35 \\
\midrule
EG-E5  & 0.66 \\
EverGreen  & 0.77 \\
\bottomrule
\end{tabular}
\caption{Correlation of ChatGPT with UC and EG. All results are significant (p-value < 0.05). EverGreen denotes ground true labels in the selected dataset part.}
\label{tab:corr_ChatGPT_UC}
\end{table}

\section{Explaining GPT-4o Retrieval} \label{sec:gpt4_retrieve}

\begin{table*}[ht!]
\centering
\small
\setlength{\tabcolsep}{2pt}
\resizebox{\textwidth}{!}{
\begin{tabular}{ll}
\toprule
\multicolumn{1}{c}{\textbf{Misclassification reason}} & \multicolumn{1}{c}{\textbf{Example questions}} \\
\midrule

\multicolumn{2}{c}{\textit{False Positives (non-evergreen, but classified as evergreen)}} \\
\midrule

\rowcolor{violet!10}
Superlatives assumed to be static facts & 
\begin{tabular}[c]{@{}l@{}} $\cdot$ What is the biggest star in the sky? \\  $\cdot$ Which tea is the healthiest? \\  $\cdot$ What is the most popular social network in the world?\end{tabular} \\

\rowcolor{violet!20}
Biographical/life data on alive people treated as static & 
\begin{tabular}[c]{@{}l@{}}  $\cdot$ In which movies has
Simu Liu
acted?\\  $\cdot$ How many works has Stephen King written?\end{tabular} \\

\midrule
\multicolumn{2}{c}{\textit{False Negatives (evergreen, but classified as non-evergreen)}} \\
\midrule

\rowcolor{green!10}
Superlatives treated as time-sensitive or trend-based & 
\begin{tabular}[c]{@{}l@{}}  $\cdot$ What is the oldest currency? \\  $\cdot$ The rarest element in the periodic table. \\  $\cdot$ How long did the shortest war in history last?\end{tabular} \\

\rowcolor{green!20}
\begin{tabular}[c]{@{}l@{}}Biological and geographical facts wrongly assumed \\ to change frequently\end{tabular} & 
\begin{tabular}[c]{@{}l@{}}  $\cdot$ What is the area of Liechtenstein? \\  $\cdot$ Which animal has the highest blood pressure?\\  $\cdot$ How many species of elephants currently live on the planet?\end{tabular} \\

\bottomrule
\end{tabular}
}
\caption{Error Analysis of EG-E5 Classifier: breakdown of misclassification patterns. }
\label{tab:error_analysis}
\end{table*}

GPT-4o autonomously decides when to invoke its retrieval system using internal, black-box criteria. We find that question evergreen-ness is the strongest predictor of this behavior, suggesting that GPT-4o's use of external search is closely linked to the temporal nature of the input.

We use the same subset, as in Section~\ref{sec:internal} -- and queried GPT-4o via its web interface, recording whether it triggered a retrieval call.

In addition to evergreen labels, we evaluated several uncertainty-based signals from Section~\ref{sec:internal} and \textbf{EG-E5} to assess their correlation with GPT-4o’s retrieval decisions.

As shown in Table~\ref{tab:corr_ChatGPT_UC}, evergreen-ness and \textbf{EG-E5} predictions are substantially stronger predictors than any uncertainty-based signal -- more than twice as informative. This suggests that GPT-4o may internally model question temporality or is guided by a retrieval policy highly sensitive to it.

\begin{takeawaybox}
Evergreen-ness is the strongest predictor of GPT-4o’s retrieval behavior, suggesting that retrieval is closely tied to temporality.
\end{takeawaybox}

\section{Error Analysis}

We selected a test part from our EverGreenQA dataset and conducted a qualitative analysis of the errors made by the EG-E5 classifier. Table~\ref{tab:error_analysis} presents examples of false positives and false negatives, grouped by cause. Notably, the classifier shows high uncertainty with superlatives — sometimes flagging them as volatile, and other times misinterpreting trend-sensitive phrases like 'most,' 'biggest,' or 'healthiest' as universally fixed. Other errors include misclassifying achievements of living people as dead and incorrectly treating stable geographical or biological facts as time-sensitive.

Interestingly, there are twice as many false negatives as false positives. This suggests that the classifier is more cautious when deciding whether a question refers to a stable fact. In some cases, external information is crucial. For example, if a person is dead, all questions about them would be evergreen, but the model needs to know whether the person is still alive. Similarly, questions about recent years (e.g., 2023–2024) pose a challenge, as the model lacks awareness of the current date. In other cases, there is room for improvement in how the model organizes and distinguishes its knowledge. For instance, learning to differentiate between truly stable physical facts (such as the area of Liechtenstein) and more variable ones (like the brightest star in the sky), or between completed historical events (e.g., the French Revolution) and ongoing developments (such as upcoming presidential elections).

Additional examples are provided in Appendix~\ref{appendix:error_analysis_detailed}.


\section*{Conclusion}

In this study, we explored the concept of evergreen-ness, whether the answer change over time. We examined the ability of LLMs to detect it and demonstrated its usefulness across several applications.

To support this investigation, we introduce \textbf{EverGreenQA}, a new multilingual dataset comprising 4,757 examples across 7 languages. Using this dataset, we benchmark modern LLMs on the task of evergreen question classification and train \textbf{EG-E5} -- a lightweight classifier that outperforms both LLMs and previously trained methods. 

We further analyze whether LLMs implicitly encode evergreen-ness through their uncertainty estimations and find that they do it a little with larger model doing it more. We further enhance existing uncertainty estimators with predicted evergreen probabilities, yielding consistent improvements.

We also show that our evergreen classifier helps curate high-quality QA datasets and supports more reliable and fair evaluations. Finally, we demonstrate that evergreenness is the best predictor of GPT-4o's search behavior, outperforming all other tested factors.

\section*{Limitations}

\begin{itemize}

\item While our {EverGreenQA} dataset is the first multilingual, human-curated benchmark for question temporality, its size remains relatively modest (3,278 examples). Nonetheless, it offers high-quality coverage across seven diverse languages and is sufficient to reveal clear trends in model behavior.

\item Although we cover 7 languages, the dataset does not span all major language families, and performance in truly low-resource settings remains unexplored. That said, our selection includes both Latin and non-Latin scripts, enabling meaningful multilingual evaluation.

\item Our LLM evaluation includes 14 models across a wide range of scales and families, but we primarily focus on representative models from each size tier. Extending to more instruction-tuned or domain-adapted variants could further generalize the findings.

\item For uncertainty-based analysis, we focus on five representative metrics. While these are widely used and sufficient to draw strong conclusions, incorporating more recent or task-specific metrics may provide additional insights.

\item Our trained evergreen classifier demonstrates strong results, but we perform only limited ablations on its architecture, training procedure, and the use of auxiliary data. Exploring more model variants or transfer learning strategies could further improve robustness.

\item Finally, while we demonstrate several practical uses of evergreen classification, we do not explore its potential in tasks such as active learning, answer calibration, or search reranking. We leave these promising directions for future work.

\end{itemize}

\section*{Ethical Considerations}

Our work involves the construction and analysis of a multilingual QA dataset, as well as the evaluation of LLM and classifier-based approaches for detecting question temporality. We made a great effort to take into account following ethical considerations and discuss them to prevent misusage:

All questions in the constructed dataset were sourced from anonymized real-user queries during internal alpha testing. No personally identifiable information (PII) was collected, stored, or used. All examples are factual in nature and were manually reviewed to ensure compliance with privacy and ethical standards.

Dataset labels and translations were created by trained linguists and multilingual annotators. Annotators were compensated fairly according to local labor regulations. We ensured that the task complexity was reasonable and the working conditions were ethical.

The lightweight classifier and dataset are intended to support research in trustworthy QA and dataset curation. We caution against deploying these tools in high-stakes applications without rigorous domain-specific validation.

Although evergreen classification can help flag outdated or unstable information, it should not be viewed as a substitute for fact verification or timeliness. We explicitly discourage the use of our tools for censorship or exclusion of mutable information inappropriately.

We believe this work contributes to more transparent and interpretable QA systems by introducing temporality as an explicit factor, while taking steps to ensure fairness, privacy, and responsible development.

\bibliography{custom}

\begin{thebibliography}{44}
\providecommand{\natexlab}[1]{#1}

\bibitem[{Abdin et~al.(2024)Abdin, Aneja, Awadalla, Awadallah, Awan, Bach, Bahree, Bakhtiari, Bao, Behl, Benhaim, Bilenko, Bjorck, Bubeck, Cai, Cai, Chaudhary, Chen, Chen, Chen, Chen, Chen, Cheng, Chopra, Dai, Dixon, Eldan, Fragoso, Gao, Gao, Gao, Garg, Giorno, Goswami, Gunasekar, Haider, Hao, Hewett, Hu, Huynh, Iter, Jacobs, Javaheripi, Jin, Karampatziakis, Kauffmann, Khademi, Kim, Kim, Kurilenko, Lee, Lee, Li, Li, Liang, Liden, Lin, Lin, Liu, Liu, Liu, Liu, Liu, Luo, Madan, Mahmoudzadeh, Majercak, Mazzola, Mendes, Mitra, Modi, Nguyen, Norick, Patra, Perez-Becker, Portet, Pryzant, Qin, Radmilac, Ren, de~Rosa, Rosset, Roy, Ruwase, Saarikivi, Saied, Salim, Santacroce, Shah, Shang, Sharma, Shen, Shukla, Song, Tanaka, Tupini, Vaddamanu, Wang, Wang, Wang, Wang, Wang, Wang, Ward, Wen, Witte, Wu, Wu, Wyatt, Xiao, Xu, Xu, Xu, Xue, Yadav, Yang, Yang, Yang, Yang, Yu, Yuan, Zhang, Zhang, Zhang, Zhang, Zhang, Zhang, Zhang, and Zhou}]{abdin2024phi3technicalreporthighly}
Marah Abdin, Jyoti Aneja, Hany Awadalla, Ahmed Awadallah, Ammar~Ahmad Awan, Nguyen Bach, Amit Bahree, Arash Bakhtiari, Jianmin Bao, Harkirat Behl, Alon Benhaim, Misha Bilenko, Johan Bjorck, Sébastien Bubeck, Martin Cai, Qin Cai, Vishrav Chaudhary, Dong Chen, Dongdong Chen, and 110 others. 2024.
\newblock \href {https://arxiv.org/abs/2404.14219} {Phi-3 technical report: A highly capable language model locally on your phone}.
\newblock \emph{Preprint}, arXiv:2404.14219.

\bibitem[{Buitinck et~al.(2013)Buitinck, Louppe, Blondel, Pedregosa, Mueller, Grisel, Niculae, Prettenhofer, Gramfort, Grobler, Layton, VanderPlas, Joly, Holt, and Varoquaux}]{sklearn_api}
Lars Buitinck, Gilles Louppe, Mathieu Blondel, Fabian Pedregosa, Andreas Mueller, Olivier Grisel, Vlad Niculae, Peter Prettenhofer, Alexandre Gramfort, Jaques Grobler, Robert Layton, Jake VanderPlas, Arnaud Joly, Brian Holt, and Ga{\"{e}}l Varoquaux. 2013.
\newblock {API} design for machine learning software: experiences from the scikit-learn project.
\newblock In \emph{ECML PKDD Workshop: Languages for Data Mining and Machine Learning}, pages 108--122.

\bibitem[{Chen et~al.(2021)Chen, Wang, and Wang}]{Chen2021ADF}
Wenhu Chen, Xinyi Wang, and William~Yang Wang. 2021.
\newblock \href {https://datasets-benchmarks-proceedings.neurips.cc/paper/2021/hash/1f0e3dad99908345f7439f8ffabdffc4-Abstract-round2.html} {A dataset for answering time-sensitive questions}.
\newblock In \emph{Proceedings of the Neural Information Processing Systems Track on Datasets and Benchmarks 1, NeurIPS Datasets and Benchmarks 2021, December 2021, virtual}.

\bibitem[{Cheng et~al.(2024)Cheng, Li, Li, Zhu, Yin, Shao, Li, Sun, Yan, and Qiu}]{cheng2024unifiedactiveretrievalretrieval}
Qinyuan Cheng, Xiaonan Li, Shimin Li, Qin Zhu, Zhangyue Yin, Yunfan Shao, Linyang Li, Tianxiang Sun, Hang Yan, and Xipeng Qiu. 2024.
\newblock \href {https://arxiv.org/abs/2406.12534} {Unified active retrieval for retrieval augmented generation}.
\newblock \emph{Preprint}, arXiv:2406.12534.

\bibitem[{Devlin et~al.(2019)Devlin, Chang, Lee, and Toutanova}]{devlin2019bertpretrainingdeepbidirectional}
Jacob Devlin, Ming-Wei Chang, Kenton Lee, and Kristina Toutanova. 2019.
\newblock \href {https://arxiv.org/abs/1810.04805} {Bert: Pre-training of deep bidirectional transformers for language understanding}.
\newblock \emph{Preprint}, arXiv:1810.04805.

\bibitem[{Ding et~al.(2024)Ding, Pang, Wei, Shen, and Cheng}]{DBLP:journals/corr/abs-2402-10612-rowen}
Hanxing Ding, Liang Pang, Zihao Wei, Huawei Shen, and Xueqi Cheng. 2024.
\newblock \href {https://doi.org/10.48550/ARXIV.2402.10612} {Retrieve only when it needs: Adaptive retrieval augmentation for hallucination mitigation in large language models}.
\newblock \emph{CoRR}, abs/2402.10612.

\bibitem[{Duan et~al.(2023)Duan, Cheng, Wang, Wang, Zavalny, Xu, Kailkhura, and Xu}]{duan2023shifting}
Jinhao Duan, Hao Cheng, Shiqi Wang, Chenan Wang, Alex Zavalny, Renjing Xu, Bhavya Kailkhura, and Kaidi Xu. 2023.
\newblock Shifting attention to relevance: Towards the uncertainty estimation of large language models.
\newblock \emph{arXiv preprint arXiv:2307.01379}.

\bibitem[{Fadeeva et~al.(2024)Fadeeva, Rubashevskii, Shelmanov, Petrakov, Li, Mubarak, Tsymbalov, Kuzmin, Panchenko, Baldwin et~al.}]{fadeeva2024fact}
Ekaterina Fadeeva, Aleksandr Rubashevskii, Artem Shelmanov, Sergey Petrakov, Haonan Li, Hamdy Mubarak, Evgenii Tsymbalov, Gleb Kuzmin, Alexander Panchenko, Timothy Baldwin, and 1 others. 2024.
\newblock Fact-checking the output of large language models via token-level uncertainty quantification.
\newblock \emph{arXiv preprint arXiv:2403.04696}.

\bibitem[{Fierro et~al.(2024)Fierro, Garneau, Bugliarello, Kementchedjhieva, and S{\o}gaard}]{DBLP:conf/naacl/FierroGBKS24}
Constanza Fierro, Nicolas Garneau, Emanuele Bugliarello, Yova Kementchedjhieva, and Anders S{\o}gaard. 2024.
\newblock \href {https://doi.org/10.18653/V1/2024.NAACL-SHORT.67} {Mulan: {A} study of fact mutability in language models}.
\newblock In \emph{Proceedings of the 2024 Conference of the North American Chapter of the Association for Computational Linguistics: Human Language Technologies: Short Papers, {NAACL} 2024, Mexico City, Mexico, June 16-21, 2024}, pages 762--771. Association for Computational Linguistics.

\bibitem[{Fomicheva et~al.(2020)Fomicheva, Sun, Yankovskaya, Blain, Guzm{\'a}n, Fishel, Aletras, Chaudhary, and Specia}]{fomicheva2020unsupervised}
Marina Fomicheva, Shuo Sun, Lisa Yankovskaya, Fr{\'e}d{\'e}ric Blain, Francisco Guzm{\'a}n, Mark Fishel, Nikolaos Aletras, Vishrav Chaudhary, and Lucia Specia. 2020.
\newblock Unsupervised quality estimation for neural machine translation.
\newblock \emph{Transactions of the Association for Computational Linguistics}, 8:539--555.

\bibitem[{Ge et~al.(2024)Ge, Mousavi, Grave, Joulin, Qian, Han, Arefiyan, and Li}]{ge-etal-2024-time}
Xiou Ge, Ali Mousavi, Edouard Grave, Armand Joulin, Kun Qian, Benjamin Han, Mostafa Arefiyan, and Yunyao Li. 2024.
\newblock \href {https://doi.org/10.18653/v1/2024.acl-short.53} {Time sensitive knowledge editing through efficient finetuning}.
\newblock In \emph{Proceedings of the 62nd Annual Meeting of the Association for Computational Linguistics (Volume 2: Short Papers)}, pages 583--593, Bangkok, Thailand. Association for Computational Linguistics.

\bibitem[{Grattafiori et~al.(2024)Grattafiori, Dubey, Jauhri, Pandey, Kadian, Al-Dahle, Letman, Mathur, Schelten, Vaughan, Yang, Fan, Goyal, Hartshorn, Yang, Mitra, Sravankumar, Korenev, Hinsvark, Rao, Zhang, Rodriguez, Gregerson, Spataru, Roziere, Biron, Tang, Chern, Caucheteux, Nayak, Bi, Marra, McConnell, Keller, Touret, Wu, Wong, Ferrer, Nikolaidis, Allonsius, Song, Pintz, Livshits, Wyatt, Esiobu, Choudhary, Mahajan, Garcia-Olano, Perino, Hupkes, Lakomkin, AlBadawy, Lobanova, Dinan, Smith, Radenovic, Guzmán, Zhang, Synnaeve, Lee, Anderson, Thattai, Nail, Mialon, Pang, Cucurell, Nguyen, Korevaar, Xu, Touvron, Zarov, Ibarra, Kloumann, Misra, Evtimov, Zhang, Copet, Lee, Geffert, Vranes, Park, Mahadeokar, Shah, van~der Linde, Billock, Hong, Lee, Fu, Chi, Huang, Liu, Wang, Yu, Bitton, Spisak, Park, Rocca, Johnstun, Saxe, Jia, Alwala, Prasad, Upasani, Plawiak, Li, Heafield, Stone, El-Arini, Iyer, Malik, Chiu, Bhalla, Lakhotia, Rantala-Yeary, van~der Maaten, Chen, Tan, Jenkins, Martin, Madaan, Malo, Blecher,
  Landzaat, de~Oliveira, Muzzi, Pasupuleti, Singh, Paluri, Kardas, Tsimpoukelli, Oldham, Rita, Pavlova, Kambadur, Lewis, Si, Singh, Hassan, Goyal, Torabi, Bashlykov, Bogoychev, Chatterji, Zhang, Duchenne, Çelebi, Alrassy, Zhang, Li, Vasic, Weng, Bhargava, Dubal, Krishnan, Koura, Xu, He, Dong, Srinivasan, Ganapathy, Calderer, Cabral, Stojnic, Raileanu, Maheswari, Girdhar, Patel, Sauvestre, Polidoro, Sumbaly, Taylor, Silva, Hou, Wang, Hosseini, Chennabasappa, Singh, Bell, Kim, Edunov, Nie, Narang, Raparthy, Shen, Wan, Bhosale, Zhang, Vandenhende, Batra, Whitman, Sootla, Collot, Gururangan, Borodinsky, Herman, Fowler, Sheasha, Georgiou, Scialom, Speckbacher, Mihaylov, Xiao, Karn, Goswami, Gupta, Ramanathan, Kerkez, Gonguet, Do, Vogeti, Albiero, Petrovic, Chu, Xiong, Fu, Meers, Martinet, Wang, Wang, Tan, Xia, Xie, Jia, Wang, Goldschlag, Gaur, Babaei, Wen, Song, Zhang, Li, Mao, Coudert, Yan, Chen, Papakipos, Singh, Srivastava, Jain, Kelsey, Shajnfeld, Gangidi, Victoria, Goldstand, Menon, Sharma, Boesenberg,
  Baevski, Feinstein, Kallet, Sangani, Teo, Yunus, Lupu, Alvarado, Caples, Gu, Ho, Poulton, Ryan, Ramchandani, Dong, Franco, Goyal, Saraf, Chowdhury, Gabriel, Bharambe, Eisenman, Yazdan, James, Maurer, Leonhardi, Huang, Loyd, Paola, Paranjape, Liu, Wu, Ni, Hancock, Wasti, Spence, Stojkovic, Gamido, Montalvo, Parker, Burton, Mejia, Liu, Wang, Kim, Zhou, Hu, Chu, Cai, Tindal, Feichtenhofer, Gao, Civin, Beaty, Kreymer, Li, Adkins, Xu, Testuggine, David, Parikh, Liskovich, Foss, Wang, Le, Holland, Dowling, Jamil, Montgomery, Presani, Hahn, Wood, Le, Brinkman, Arcaute, Dunbar, Smothers, Sun, Kreuk, Tian, Kokkinos, Ozgenel, Caggioni, Kanayet, Seide, Florez, Schwarz, Badeer, Swee, Halpern, Herman, Sizov, Guangyi, Zhang, Lakshminarayanan, Inan, Shojanazeri, Zou, Wang, Zha, Habeeb, Rudolph, Suk, Aspegren, Goldman, Zhan, Damlaj, Molybog, Tufanov, Leontiadis, Veliche, Gat, Weissman, Geboski, Kohli, Lam, Asher, Gaya, Marcus, Tang, Chan, Zhen, Reizenstein, Teboul, Zhong, Jin, Yang, Cummings, Carvill, Shepard, McPhie,
  Torres, Ginsburg, Wang, Wu, U, Saxena, Khandelwal, Zand, Matosich, Veeraraghavan, Michelena, Li, Jagadeesh, Huang, Chawla, Huang, Chen, Garg, A, Silva, Bell, Zhang, Guo, Yu, Moshkovich, Wehrstedt, Khabsa, Avalani, Bhatt, Mankus, Hasson, Lennie, Reso, Groshev, Naumov, Lathi, Keneally, Liu, Seltzer, Valko, Restrepo, Patel, Vyatskov, Samvelyan, Clark, Macey, Wang, Hermoso, Metanat, Rastegari, Bansal, Santhanam, Parks, White, Bawa, Singhal, Egebo, Usunier, Mehta, Laptev, Dong, Cheng, Chernoguz, Hart, Salpekar, Kalinli, Kent, Parekh, Saab, Balaji, Rittner, Bontrager, Roux, Dollar, Zvyagina, Ratanchandani, Yuvraj, Liang, Alao, Rodriguez, Ayub, Murthy, Nayani, Mitra, Parthasarathy, Li, Hogan, Battey, Wang, Howes, Rinott, Mehta, Siby, Bondu, Datta, Chugh, Hunt, Dhillon, Sidorov, Pan, Mahajan, Verma, Yamamoto, Ramaswamy, Lindsay, Lindsay, Feng, Lin, Zha, Patil, Shankar, Zhang, Zhang, Wang, Agarwal, Sajuyigbe, Chintala, Max, Chen, Kehoe, Satterfield, Govindaprasad, Gupta, Deng, Cho, Virk, Subramanian, Choudhury,
  Goldman, Remez, Glaser, Best, Koehler, Robinson, Li, Zhang, Matthews, Chou, Shaked, Vontimitta, Ajayi, Montanez, Mohan, Kumar, Mangla, Ionescu, Poenaru, Mihailescu, Ivanov, Li, Wang, Jiang, Bouaziz, Constable, Tang, Wu, Wang, Wu, Gao, Kleinman, Chen, Hu, Jia, Qi, Li, Zhang, Zhang, Adi, Nam, Yu, Wang, Zhao, Hao, Qian, Li, He, Rait, DeVito, Rosnbrick, Wen, Yang, Zhao, and Ma}]{grattafiori2024llama3herdmodels}
Aaron Grattafiori, Abhimanyu Dubey, Abhinav Jauhri, Abhinav Pandey, Abhishek Kadian, Ahmad Al-Dahle, Aiesha Letman, Akhil Mathur, Alan Schelten, Alex Vaughan, Amy Yang, Angela Fan, Anirudh Goyal, Anthony Hartshorn, Aobo Yang, Archi Mitra, Archie Sravankumar, Artem Korenev, Arthur Hinsvark, and 542 others. 2024.
\newblock \href {https://arxiv.org/abs/2407.21783} {The llama 3 herd of models}.
\newblock \emph{Preprint}, arXiv:2407.21783.

\bibitem[{Hancock and Khoshgoftaar(2020)}]{hancock2020catboost}
John~T Hancock and Taghi~M Khoshgoftaar. 2020.
\newblock Catboost for big data: an interdisciplinary review.
\newblock \emph{Journal of big data}, 7(1):94.

\bibitem[{He et~al.(2023)He, Gao, and Chen}]{he2023debertav3improvingdebertausing}
Pengcheng He, Jianfeng Gao, and Weizhu Chen. 2023.
\newblock \href {https://arxiv.org/abs/2111.09543} {Debertav3: Improving deberta using electra-style pre-training with gradient-disentangled embedding sharing}.
\newblock \emph{Preprint}, arXiv:2111.09543.

\bibitem[{Ho et~al.(2020)Ho, Nguyen, Sugawara, and Aizawa}]{DBLP:conf/coling/HoNSA20}
Xanh Ho, Anh{-}Khoa~Duong Nguyen, Saku Sugawara, and Akiko Aizawa. 2020.
\newblock \href {https://doi.org/10.18653/V1/2020.COLING-MAIN.580} {Constructing {A} multi-hop {QA} dataset for comprehensive evaluation of reasoning steps}.
\newblock In \emph{Proceedings of the 28th International Conference on Computational Linguistics, {COLING} 2020, Barcelona, Spain (Online), December 8-13, 2020}, pages 6609--6625. International Committee on Computational Linguistics.

\bibitem[{Huang et~al.(2025)Huang, Yu, Ma, Zhong, Feng, Wang, Chen, Peng, Feng, Qin, and Liu}]{Huang_2025}
Lei Huang, Weijiang Yu, Weitao Ma, Weihong Zhong, Zhangyin Feng, Haotian Wang, Qianglong Chen, Weihua Peng, Xiaocheng Feng, Bing Qin, and Ting Liu. 2025.
\newblock \href {https://doi.org/10.1145/3703155} {A survey on hallucination in large language models: Principles, taxonomy, challenges, and open questions}.
\newblock \emph{ACM Transactions on Information Systems}, 43(2):1–55.

\bibitem[{Jeong et~al.(2024)Jeong, Baek, Cho, Hwang, and Park}]{DBLP:conf/naacl/JeongBCHP24}
Soyeong Jeong, Jinheon Baek, Sukmin Cho, Sung~Ju Hwang, and Jong Park. 2024.
\newblock \href {https://doi.org/10.18653/V1/2024.NAACL-LONG.389} {Adaptive-rag: Learning to adapt retrieval-augmented large language models through question complexity}.
\newblock In \emph{Proceedings of the 2024 Conference of the North American Chapter of the Association for Computational Linguistics: Human Language Technologies (Volume 1: Long Papers), {NAACL} 2024, Mexico City, Mexico, June 16-21, 2024}, pages 7036--7050. Association for Computational Linguistics.

\bibitem[{Jiang et~al.(2023)Jiang, Sablayrolles, Mensch, Bamford, Chaplot, de~las Casas, Bressand, Lengyel, Lample, Saulnier, Lavaud, Lachaux, Stock, Scao, Lavril, Wang, Lacroix, and Sayed}]{jiang2023mistral7b}
Albert~Q. Jiang, Alexandre Sablayrolles, Arthur Mensch, Chris Bamford, Devendra~Singh Chaplot, Diego de~las Casas, Florian Bressand, Gianna Lengyel, Guillaume Lample, Lucile Saulnier, Lélio~Renard Lavaud, Marie-Anne Lachaux, Pierre Stock, Teven~Le Scao, Thibaut Lavril, Thomas Wang, Timothée Lacroix, and William~El Sayed. 2023.
\newblock \href {https://arxiv.org/abs/2310.06825} {Mistral 7b}.
\newblock \emph{Preprint}, arXiv:2310.06825.

\bibitem[{Joshi et~al.(2017)Joshi, Choi, Weld, and Zettlemoyer}]{DBLP:conf/acl/JoshiCWZ17}
Mandar Joshi, Eunsol Choi, Daniel~S. Weld, and Luke Zettlemoyer. 2017.
\newblock \href {https://doi.org/10.18653/V1/P17-1147} {Triviaqa: {A} large scale distantly supervised challenge dataset for reading comprehension}.
\newblock In \emph{Proceedings of the 55th Annual Meeting of the Association for Computational Linguistics, {ACL} 2017, Vancouver, Canada, July 30 - August 4, Volume 1: Long Papers}, pages 1601--1611. Association for Computational Linguistics.

\bibitem[{Kasai et~al.(2024)Kasai, Sakaguchi, Takahashi, Bras, Asai, Yu, Radev, Smith, Choi, and Inui}]{kasai2024realtimeqawhatsanswer}
Jungo Kasai, Keisuke Sakaguchi, Yoichi Takahashi, Ronan~Le Bras, Akari Asai, Xinyan Yu, Dragomir Radev, Noah~A. Smith, Yejin Choi, and Kentaro Inui. 2024.
\newblock \href {https://arxiv.org/abs/2207.13332} {Realtime qa: What's the answer right now?}
\newblock \emph{Preprint}, arXiv:2207.13332.

\bibitem[{Kwiatkowski et~al.(2019)Kwiatkowski, Palomaki, Redfield, Collins, Parikh, Alberti, Epstein, Polosukhin, Devlin, Lee, Toutanova, Jones, Kelcey, Chang, Dai, Uszkoreit, Le, and Petrov}]{DBLP:journals/tacl/KwiatkowskiPRCP19}
Tom Kwiatkowski, Jennimaria Palomaki, Olivia Redfield, Michael Collins, Ankur~P. Parikh, Chris Alberti, Danielle Epstein, Illia Polosukhin, Jacob Devlin, Kenton Lee, Kristina Toutanova, Llion Jones, Matthew Kelcey, Ming{-}Wei Chang, Andrew~M. Dai, Jakob Uszkoreit, Quoc Le, and Slav Petrov. 2019.
\newblock \href {https://doi.org/10.1162/TACL\_A\_00276} {Natural questions: a benchmark for question answering research}.
\newblock \emph{Trans. Assoc. Comput. Linguistics}, 7:452--466.

\bibitem[{Lin et~al.(2023)Lin, Trivedi, and Sun}]{lin2023generating}
Zhen Lin, Shubhendu Trivedi, and Jimeng Sun. 2023.
\newblock Generating with confidence: Uncertainty quantification for black-box large language models.
\newblock \emph{arXiv preprint arXiv:2305.19187}.

\bibitem[{Liska et~al.(2022)Liska, Kocisky, Gribovskaya, Terzi, Sezener, Agrawal, De~Masson~D'Autume, Scholtes, Zaheer, Young, Gilsenan-Mcmahon, Austin, Blunsom, and Lazaridou}]{pmlr-v162-liska22a}
Adam Liska, Tomas Kocisky, Elena Gribovskaya, Tayfun Terzi, Eren Sezener, Devang Agrawal, Cyprien De~Masson~D'Autume, Tim Scholtes, Manzil Zaheer, Susannah Young, Ellen Gilsenan-Mcmahon, Sophia Austin, Phil Blunsom, and Angeliki Lazaridou. 2022.
\newblock \href {https://proceedings.mlr.press/v162/liska22a.html} {{S}treaming{QA}: A benchmark for adaptation to new knowledge over time in question answering models}.
\newblock In \emph{Proceedings of the 39th International Conference on Machine Learning}, volume 162 of \emph{Proceedings of Machine Learning Research}, pages 13604--13622. PMLR.

\bibitem[{Moskvoretskii et~al.(2025)Moskvoretskii, Lysyuk, Salnikov, Ivanov, Pletenev, Galimzianova, Krayko, Konovalov, Nikishina, and Panchenko}]{moskvoretskii2025adaptive}
Viktor Moskvoretskii, Maria Lysyuk, Mikhail Salnikov, Nikolay Ivanov, Sergey Pletenev, Daria Galimzianova, Nikita Krayko, Vasily Konovalov, Irina Nikishina, and Alexander Panchenko. 2025.
\newblock Adaptive retrieval without self-knowledge? bringing uncertainty back home.
\newblock \emph{arXiv preprint arXiv:2501.12835}.

\bibitem[{Pletenev et~al.(2025)Pletenev, Marina, Moskovskiy, Konovalov, Braslavski, Panchenko, and Salnikov}]{pletenev-etal-2025-much}
Sergey Pletenev, Maria Marina, Daniil Moskovskiy, Vasily Konovalov, Pavel Braslavski, Alexander Panchenko, and Mikhail Salnikov. 2025.
\newblock \href {https://aclanthology.org/2025.findings-naacl.243/} {How much knowledge can you pack into a {L}o{RA} adapter without harming {LLM}?}
\newblock In \emph{Findings of the Association for Computational Linguistics: NAACL 2025}, pages 4309--4322, Albuquerque, New Mexico. Association for Computational Linguistics.

\bibitem[{Rajpurkar et~al.(2016)Rajpurkar, Zhang, Lopyrev, and Liang}]{DBLP:conf/emnlp/RajpurkarZLL16}
Pranav Rajpurkar, Jian Zhang, Konstantin Lopyrev, and Percy Liang. 2016.
\newblock \href {https://doi.org/10.18653/V1/D16-1264} {Squad: 100, 000+ questions for machine comprehension of text}.
\newblock In \emph{Proceedings of the 2016 Conference on Empirical Methods in Natural Language Processing, {EMNLP} 2016, Austin, Texas, USA, November 1-4, 2016}, pages 2383--2392. The Association for Computational Linguistics.

\bibitem[{Su et~al.(2024)Su, Tang, Ai, Wu, and Liu}]{DBLP:conf/acl/SuTA0024}
Weihang Su, Yichen Tang, Qingyao Ai, Zhijing Wu, and Yiqun Liu. 2024.
\newblock \href {https://doi.org/10.18653/V1/2024.ACL-LONG.702} {{DRAGIN:} dynamic retrieval augmented generation based on the real-time information needs of large language models}.
\newblock In \emph{Proceedings of the 62nd Annual Meeting of the Association for Computational Linguistics (Volume 1: Long Papers), {ACL} 2024, Bangkok, Thailand, August 11-16, 2024}, pages 12991--13013. Association for Computational Linguistics.

\bibitem[{Team et~al.(2024)Team, Riviere, Pathak, Sessa, Hardin, Bhupatiraju, Hussenot, Mesnard, Shahriari, Ramé, Ferret, Liu, Tafti, Friesen, Casbon, Ramos, Kumar, Lan, Jerome, Tsitsulin, Vieillard, Stanczyk, Girgin, Momchev, Hoffman, Thakoor, Grill, Neyshabur, Bachem, Walton, Severyn, Parrish, Ahmad, Hutchison, Abdagic, Carl, Shen, Brock, Coenen, Laforge, Paterson, Bastian, Piot, Wu, Royal, Chen, Kumar, Perry, Welty, Choquette-Choo, Sinopalnikov, Weinberger, Vijaykumar, Rogozińska, Herbison, Bandy, Wang, Noland, Moreira, Senter, Eltyshev, Visin, Rasskin, Wei, Cameron, Martins, Hashemi, Klimczak-Plucińska, Batra, Dhand, Nardini, Mein, Zhou, Svensson, Stanway, Chan, Zhou, Carrasqueira, Iljazi, Becker, Fernandez, van Amersfoort, Gordon, Lipschultz, Newlan, yeong Ji, Mohamed, Badola, Black, Millican, McDonell, Nguyen, Sodhia, Greene, Sjoesund, Usui, Sifre, Heuermann, Lago, McNealus, Soares, Kilpatrick, Dixon, Martins, Reid, Singh, Iverson, Görner, Velloso, Wirth, Davidow, Miller, Rahtz, Watson, Risdal,
  Kazemi, Moynihan, Zhang, Kahng, Park, Rahman, Khatwani, Dao, Bardoliwalla, Devanathan, Dumai, Chauhan, Wahltinez, Botarda, Barnes, Barham, Michel, Jin, Georgiev, Culliton, Kuppala, Comanescu, Merhej, Jana, Rokni, Agarwal, Mullins, Saadat, Carthy, Cogan, Perrin, Arnold, Krause, Dai, Garg, Sheth, Ronstrom, Chan, Jordan, Yu, Eccles, Hennigan, Kocisky, Doshi, Jain, Yadav, Meshram, Dharmadhikari, Barkley, Wei, Ye, Han, Kwon, Xu, Shen, Gong, Wei, Cotruta, Kirk, Rao, Giang, Peran, Warkentin, Collins, Barral, Ghahramani, Hadsell, Sculley, Banks, Dragan, Petrov, Vinyals, Dean, Hassabis, Kavukcuoglu, Farabet, Buchatskaya, Borgeaud, Fiedel, Joulin, Kenealy, Dadashi, and Andreev}]{gemmateam2024gemma2improvingopen}
Gemma Team, Morgane Riviere, Shreya Pathak, Pier~Giuseppe Sessa, Cassidy Hardin, Surya Bhupatiraju, Léonard Hussenot, Thomas Mesnard, Bobak Shahriari, Alexandre Ramé, Johan Ferret, Peter Liu, Pouya Tafti, Abe Friesen, Michelle Casbon, Sabela Ramos, Ravin Kumar, Charline~Le Lan, Sammy Jerome, and 179 others. 2024.
\newblock \href {https://arxiv.org/abs/2408.00118} {Gemma 2: Improving open language models at a practical size}.
\newblock \emph{Preprint}, arXiv:2408.00118.

\bibitem[{Team(2024)}]{qwen2.5}
Qwen Team. 2024.
\newblock \href {https://qwenlm.github.io/blog/qwen2.5/} {Qwen2.5: A party of foundation models}.

\bibitem[{Trivedi et~al.(2022)Trivedi, Balasubramanian, Khot, and Sabharwal}]{DBLP:journals/tacl/TrivediBKS22}
Harsh Trivedi, Niranjan Balasubramanian, Tushar Khot, and Ashish Sabharwal. 2022.
\newblock \href {https://doi.org/10.1162/TACL\_A\_00475} {Musique: Multihop questions via single-hop question composition}.
\newblock \emph{Trans. Assoc. Comput. Linguistics}, 10:539--554.

\bibitem[{Trivedi et~al.(2023)Trivedi, Balasubramanian, Khot, and Sabharwal}]{DBLP:conf/acl/TrivediBKS23}
Harsh Trivedi, Niranjan Balasubramanian, Tushar Khot, and Ashish Sabharwal. 2023.
\newblock \href {https://doi.org/10.18653/V1/2023.ACL-LONG.557} {Interleaving retrieval with chain-of-thought reasoning for knowledge-intensive multi-step questions}.
\newblock In \emph{Proceedings of the 61st Annual Meeting of the Association for Computational Linguistics (Volume 1: Long Papers), {ACL} 2023, Toronto, Canada, July 9-14, 2023}, pages 10014--10037. Association for Computational Linguistics.

\bibitem[{Vashurin et~al.(2024)Vashurin, Fadeeva, Vazhentsev, Rvanova, Vasilev, Tsvigun, Petrakov, Xing, Sadallah, Grishchenkov, Panchenko, Baldwin, Nakov, Panov, and Shelmanov}]{10.1162/tacl_a_00737}
Roman Vashurin, Ekaterina Fadeeva, Artem Vazhentsev, Lyudmila Rvanova, Daniil Vasilev, Akim Tsvigun, Sergey Petrakov, Rui Xing, Abdelrahman Sadallah, Kirill Grishchenkov, Alexander Panchenko, Timothy Baldwin, Preslav Nakov, Maxim Panov, and Artem Shelmanov. 2024.
\newblock \href {https://doi.org/10.1162/tacl_a_00737} {Benchmarking uncertainty quantification methods for large language models with lm-polygraph}.
\newblock \emph{Transactions of the Association for Computational Linguistics}, 13:220--248.

\bibitem[{Vayani et~al.(2024)Vayani, Dissanayake, Watawana, Ahsan, Sasikumar, Thawakar, Ademtew, Hmaiti, Kumar, Kuckreja et~al.}]{vayani2024all}
Ashmal Vayani, Dinura Dissanayake, Hasindri Watawana, Noor Ahsan, Nevasini Sasikumar, Omkar Thawakar, Henok~Biadglign Ademtew, Yahya Hmaiti, Amandeep Kumar, Kartik Kuckreja, and 1 others. 2024.
\newblock All languages matter: Evaluating lmms on culturally diverse 100 languages.
\newblock \emph{arXiv preprint arXiv:2411.16508}.

\bibitem[{Vazhentsev et~al.(2025)Vazhentsev, Rvanova, Lazichny, Panchenko, Panov, Baldwin, and Shelmanov}]{vazhentsev2025token}
Artem Vazhentsev, Lyudmila Rvanova, Ivan Lazichny, Alexander Panchenko, Maxim Panov, Timothy Baldwin, and Artem Shelmanov. 2025.
\newblock Token-level density-based uncertainty quantification methods for eliciting truthfulness of large language models.
\newblock \emph{arXiv preprint arXiv:2502.14427}.

\bibitem[{Veldanda et~al.(2024)Veldanda, Zhang, Das, Chakraborty, Rawls, Sahu, and Naphade}]{veldanda2024llm}
Akshaj~Kumar Veldanda, Shi-Xiong Zhang, Anirban Das, Supriyo Chakraborty, Stephen Rawls, Sambit Sahu, and Milind Naphade. 2024.
\newblock Llm surgery: Efficient knowledge unlearning and editing in large language models.
\newblock \emph{arXiv e-prints}, pages arXiv--2409.

\bibitem[{Vu et~al.(2024)Vu, Iyyer, Wang, Constant, Wei, Wei, Tar, Sung, Zhou, Le, and Luong}]{DBLP:conf/acl/VuI0CWWTSZLL24}
Tu~Vu, Mohit Iyyer, Xuezhi Wang, Noah Constant, Jerry~W. Wei, Jason Wei, Chris Tar, Yun{-}Hsuan Sung, Denny Zhou, Quoc~V. Le, and Thang Luong. 2024.
\newblock \href {https://doi.org/10.18653/V1/2024.FINDINGS-ACL.813} {Freshllms: Refreshing large language models with search engine augmentation}.
\newblock In \emph{Findings of the Association for Computational Linguistics, {ACL} 2024, Bangkok, Thailand and virtual meeting, August 11-16, 2024}, pages 13697--13720. Association for Computational Linguistics.

\bibitem[{Wang et~al.(2024)Wang, Yang, Huang, Yang, Majumder, and Wei}]{wang2024multilinguale5textembeddings}
Liang Wang, Nan Yang, Xiaolong Huang, Linjun Yang, Rangan Majumder, and Furu Wei. 2024.
\newblock \href {https://arxiv.org/abs/2402.05672} {Multilingual e5 text embeddings: A technical report}.
\newblock \emph{Preprint}, arXiv:2402.05672.

\bibitem[{Wang et~al.(2023)Wang, Li, Sun, and Liu}]{DBLP:conf/emnlp/WangLSL23}
Yile Wang, Peng Li, Maosong Sun, and Yang Liu. 2023.
\newblock \href {https://doi.org/10.18653/V1/2023.FINDINGS-EMNLP.691} {Self-knowledge guided retrieval augmentation for large language models}.
\newblock In \emph{Findings of the Association for Computational Linguistics: {EMNLP} 2023, Singapore, December 6-10, 2023}, pages 10303--10315. Association for Computational Linguistics.

\bibitem[{Wei et~al.(2024{\natexlab{a}})Wei, Karina, Chung, Jiao, Papay, Glaese, Schulman, and Fedus}]{wei2024measuringshortformfactualitylarge}
Jason Wei, Nguyen Karina, Hyung~Won Chung, Yunxin~Joy Jiao, Spencer Papay, Amelia Glaese, John Schulman, and William Fedus. 2024{\natexlab{a}}.
\newblock \href {https://arxiv.org/abs/2411.04368} {Measuring short-form factuality in large language models}.
\newblock \emph{Preprint}, arXiv:2411.04368.

\bibitem[{Wei et~al.(2024{\natexlab{b}})Wei, Karina, Chung, Jiao, Papay, Glaese, Schulman, and Fedus}]{wei2024measuring}
Jason Wei, Nguyen Karina, Hyung~Won Chung, Yunxin~Joy Jiao, Spencer Papay, Amelia Glaese, John Schulman, and William Fedus. 2024{\natexlab{b}}.
\newblock Measuring short-form factuality in large language models.
\newblock \emph{arXiv preprint arXiv:2411.04368}.

\bibitem[{Yang et~al.(2018)Yang, Qi, Zhang, Bengio, Cohen, Salakhutdinov, and Manning}]{DBLP:conf/emnlp/Yang0ZBCSM18}
Zhilin Yang, Peng Qi, Saizheng Zhang, Yoshua Bengio, William~W. Cohen, Ruslan Salakhutdinov, and Christopher~D. Manning. 2018.
\newblock \href {https://doi.org/10.18653/V1/D18-1259} {Hotpotqa: {A} dataset for diverse, explainable multi-hop question answering}.
\newblock In \emph{Proceedings of the 2018 Conference on Empirical Methods in Natural Language Processing, Brussels, Belgium, October 31 - November 4, 2018}, pages 2369--2380. Association for Computational Linguistics.

\bibitem[{Yin et~al.(2023)Yin, Sun, Guo, Wu, Qiu, and Huang}]{yin2023large}
Zhangyue Yin, Qiushi Sun, Qipeng Guo, Jiawen Wu, Xipeng Qiu, and Xuanjing Huang. 2023.
\newblock Do large language models know what they don't know?
\newblock \emph{arXiv preprint arXiv:2305.18153}.

\bibitem[{Zhang and Choi(2021)}]{DBLP:conf/emnlp/ZhangC21}
Michael J.~Q. Zhang and Eunsol Choi. 2021.
\newblock \href {https://doi.org/10.18653/V1/2021.EMNLP-MAIN.586} {Situatedqa: Incorporating extra-linguistic contexts into {QA}}.
\newblock In \emph{Proceedings of the 2021 Conference on Empirical Methods in Natural Language Processing, {EMNLP} 2021, Virtual Event / Punta Cana, Dominican Republic, 7-11 November, 2021}, pages 7371--7387. Association for Computational Linguistics.

\bibitem[{Zhao et~al.(2024)Zhao, Brumbaugh, Wang, Hajishirzi, and Smith}]{DBLP:conf/acl/ZhaoBWHS24}
Bowen Zhao, Zander Brumbaugh, Yizhong Wang, Hannaneh Hajishirzi, and Noah~A. Smith. 2024.
\newblock \href {https://doi.org/10.18653/V1/2024.FINDINGS-ACL.892} {Set the clock: Temporal alignment of pretrained language models}.
\newblock In \emph{Findings of the Association for Computational Linguistics, {ACL} 2024, Bangkok, Thailand and virtual meeting, August 11-16, 2024}, pages 15015--15040. Association for Computational Linguistics.

\end{thebibliography}

\appendix

\section{Evergreen Testing Details} \label{appendix:model_evergreen}

\textbf{LLM Verbal Parameters.} Each example comes with 5-shot for mutable and 5-shot for immutable examples. For llama 3.1 sampling parameters are following: \textsc{temperature=0.7, top\_p=0.9}. For Qwen 2.5: \textsc{temperature=0.6, top\_p=0.95, top\_k=20, min\_p=0}

\textbf{Our Classificator Parameters.} All models were trained for 10 epochs with early-stopping and \texttt{lr = 4.6e-5, bs = 16}. Additional datasets were not used. We trained one model for all languages. As shown in Table~\ref{tab:train_e5} \textbf{multilingual-e5-large-instruct} gives best results. 

\begin{tcolorbox}[colback=gray!5!white, colframe=gray!75!black, title=Evergreen Verbal Instruction]
You are a helpful assistant. You help user to classify the questions based on the temporality. There are two classes: immutable and mutable. Immutable, in which the answer almost never changes. Mutable, in which the answer typically changes over the course of several years or less. Think about each question and in the end answer with Mutable or Immutable starting with 'Classification:'
\end{tcolorbox}

\section{Translation Prompt} \label{appendix:translation_prompt}

\begin{tcolorbox}[colback=gray!5!white, colframe=gray!75!black, title=Translation Validation Instruction]
Translate the following English text into French, German, Hebrew, Arabic and Chinese. Provide the translations as a JSON object with keys 'French', 'German', 'Hebrew', 'Arabic', 'Chinese'.
\end{tcolorbox}

We use GPT 4.1 with \textsc{temperature=0.2} and additional tag \textsc{"response\_format": "json\_object"}

\section{Validation Instructions} \label{appendix:validation}

\begin{tcolorbox}[colback=gray!5!white, colframe=gray!75!black, title=Translation Validation Instruction]
For each translated question, assign a score according to the following criteria:
\begin{itemize}
    \item \textbf{0} -- the translation contains errors that \textit{distort the meaning}.
    \item \textbf{1} -- the translation contains \textit{minor errors} that do \textit{not affect the overall meaning}.
\end{itemize}
\end{tcolorbox}




\section{Classifier for Self-Knowledge} \label{appendix:training_classifier}

We explored seven classification models using \texttt{scikit-learn}~\cite{sklearn_api} and \texttt{CatBoost}~\cite{hancock2020catboost}: Logistic Regression, k-Nearest Neighbors, Multi-layer Perceptron, Decision Tree, Random Forest, Gradient Boosting, and CatBoost. All models were trained with standardized features using \texttt{StandardScaler}. Hyperparameters were optimized on a validation subset of 100 examples randomly sampled from the training data, and experiments were repeated with three random seeds per dataset to ensure robustness.

For final evaluation, we selected the two best-performing models on the validation set and combined them into a soft-voting ensemble using \texttt{VotingClassifier}. Each component model was retrained on the full training set with its tuned hyperparameters.

\paragraph{Hyperparameters grid.}

\textit{Logistic Regression} :  {C: [0.01, 0.1, 1], solver: [lbfgs, liblinear],
class\_weight: [balanced,  {0: 1, 1: 1}, None], max\_iter: [10000, 15000, 20000]}

\textit{KNN} : {n\_neighbors:  [5, 7, 9, 11, 13, 15], metric: [euclidean, manhattan], algorithm: [auto, ball\_tree, kd\_tree], weights: [uniform, distance]}

\textit{MLP} : {hidden\_layer\_sizes: [(50,), (100,), (50, 50), (100, 50), (100, 100)], activation: [relu, tanh],
solver: [adam, sgd], alpha: [0.00001, 0.0001, 0.001, 0.01],
learning\_rate: [constant, adaptive],
early\_stopping: True,
max\_iter: [200, 500]}

\textit{Decision Tree} : {max\_depth: [3, 5, 7, 10, None], max\_features: [0.2, 0.4, sqrt, log2, None],
criterion: [gini, entropy],
splitter: [best, random]}

\textit{CatBoosting}:  {iterations: [10, 50, 100, 200],
learning\_rate: [0.001, 0.01, 0.05],
depth: [3, 4, 5, 7, 9],
bootstrap\_type: [Bayesian, Bernoulli, MVS]}

\textit{Gradient Boosting}: {
n\_estimators: [25, 35, 50],
learning\_rate: [0.001, 0.01, 0.05],
max\_depth: [3, 4, 5, 7, 9],
max\_features: [0.2, 0.4, sqrt, log2, None]}

\textit{Random Forest}: {
n\_estimators: [25, 35, 50],
max\_depth: [3, 5, 7, 9, 11],
max\_features: [0.2, 0.4, sqrt, log2, None],
bootstrap: [True, False],
criterion: [gini, entropy],
class\_weight: [balanced, {0: 1, 1: 1}, None]}

\section{Predictive Analysis of Uncertainty for Temporality} \label{appendix:r2_ue}

\begin{table}[t!]
\centering
\resizebox{0.48\textwidth}{!}{
\begin{tabular}{lcc}
\toprule
{\textbf{Model}} & {\textbf{Perplexity}} & {\textbf{Mean Token Entropy}} \\
\midrule
Gemma 2 9B           & 0.014 & 0.070 \\
Gemma 2 27B          & 0.070 & 0.013 \\
LLaMA 3.1 8B         & 0.054 & 0.066 \\
LLaMA 3.1 70B        & 0.028 & 0.021 \\
Mistral 7B          & 0.016 & 0.012 \\
Mistral 24B          & 0.046 & 0.026 \\
Phi-3-mini 4k        & 0.032 & 0.016 \\
Phi-3-mini 128k      & 0.137 & 0.073 \\
Qwen 2.5 7B          & 0.025 & 0.016 \\
Qwen 2.5 32B         & 0.020 & 0.027 \\
Qwen 2.5 72B         & 0.031 & 0.029 \\
\bottomrule
\end{tabular}
}
\caption{McFadden’s pseudo-$R^2$ scores from logistic regression models trained to predict evergreen probability from two uncertainty metrics: perplexity and mean token entropy.}
\label{tab:R2_EG_UC}
\end{table}

Table~\ref{tab:R2_EG_UC} reports McFadden’s pseudo-$R^2$ values from logistic regression models trained to predict evergreen-ness based on two uncertainty metrics: perplexity and mean token entropy. 

Across most models, the pseudo-$R^2$ scores remain below $0.07$, indicating that uncertainty alone provides limited predictive power for evergreen classification. The only notable exception is {Phi-3-medium (128k)}, which achieves the highest scores--$0.137$ (perplexity)--suggesting that longer context training may improve temporal uncertainty encoding, however still very limited.

We observe no consistent advantage of one uncertainty metric over the other. Similarly, model size does not correlate clearly with predictive performance; smaller models sometimes match or outperform their larger counterparts. 

The results indicate that uncertainty metrics capture limited signals of temporality, supporting their use as complementary features rather than standalone predictors of evergreen-ness.

\section{Dataset collection details} \label{appendix:data_collection_details}

The team of trained linguists responsible for assigning the evergreen and mutable labels, as well as writing the golden answers, each hold at least a bachelor's degree in linguistics, ensuring a strong foundation in linguistic principles and effective communication. Additionally, each stage of the labeling process was carefully validated through consultation with the team lead, who provided oversight to maintain consistency and accuracy across the dataset. Furthermore, to support diverse applications, all answers were converted into a set of aliases. The procedure for this conversion is detailed in Appendix~\ref{appendix:short_answer_prompt}. The assessors were fairly paid according to local regulations.

\subsection{Golden Answers Annotation}
\label{appendix:golden_annotation}

Golden answers should be complete and useful for the user.

Examples of good and informative answers:
\textbf{Question:} Who is considered the founder of physics?
\textbf{Answer:} Isaac Newton is widely regarded as the founder of physics.
\textbf{Comment:} The question is asked in the singular form, and according to many sources, Newton is indeed considered the founder of classical physics. Based on logic, online sources, and answers from competing systems, it’s clear that Galileo Galilei and René Descartes also made significant contributions. However, since the question refers to a single person and sources support it, Newton is the most accurate and accepted answer in this context.

\textbf{Question:} Who was the President of Italy in the year 2000?
\textbf{Answer:} Carlo Azeglio Ciampi was an Italian statesman, the 10th President of the Italian Republic, and former Prime Minister of Italy.
\textbf{Comment:} A quick fact-check (as should be done for all examples in the guidelines) confirms this answer is accurate and complete.

Example of an incomplete or partially useful answer that is not suitable as a golden answer:
\textbf{Question:} Do spiders have teeth?
\textbf{Answer:}  Yes, spiders have teeth.
\textbf{Comment:} A fact-check in open sources reveals that this answer is not accurate enough to be considered a golden answer. The correct response would be: ``Spiders do not have teeth, but they have chelicerae, which contain ducts from venom glands that secrete digestive enzymes.'' Sometimes, chelicerae are colloquially referred to as ``fangs'' or ``teeth'', but they are not actually teeth. Therefore, the original answer should be revised to meet the standard of a golden answer.

\textbf{Birthday-related questions:}
If the question is phrased like \textit{How old is Yann LeCun?}, the answer should include the exact age, not just the date or year of birth.

\textbf{Open-ended list questions:}
For questions such as \textit{What are the tallest mountains?}, \textit{Which astronauts are there?}, or \textit{What animals live in Africa?}, a good answer should list at least several correct examples and include a note that this is not an exhaustive list - more exist.

\subsection{Evergreen-ness Annotation}
\label{appendix:evergreenness_annotation}

The \texttt{evergreen} criterion is a nuanced one. Most questions are considered evergreen because they are related to established facts or events. However, there are domains, such as astronomy, where new discoveries occur regularly. For example, the record for the largest known star has changed quite recently.

The definition of this criterion depends on the domain of the question. In most cases, facts that have remained unchanged for 20–30 years are treated as established. Obviously, questions like \textit{Who is the president?} are not considered evergreen due to frequent changes in political leadership.

During annotation, when we encountered ambiguous cases, we often relied on domain-specific common sense. For example, it is fairly obvious that most major geographical discoveries have already been made. It is highly unlikely that a new largest lake or a previously unknown landmass on our planet will be discovered.

As for questions involving dates, events, and notable personalities, the vast majority of these are considered evergreen, it is nearly impossible to imagine a scenario in which the dates of significant historical events or key facts from someone's biography would change.

\textbf{Mutable questions:}

(1) \textit{What year was the last solar eclipse?}

(2) \textit{Which country has the longest railway?}

(3) \textit{What date does Ramadan begin?}

(4) \textit{When is the next Olympics?}

(5) \textit{How old is Mike Tyson?}

\textbf{Evergreen questions:}

(1) \textit{Into which two states was the Roman Empire divided, and when?}

(2) \textit{Who is Messi?}

(3) \textit{Name the years of Paul von Hindenburg’s leadership in Germany.}

(4) \textit{Name the largest lakes on our planet.}

(5) \textit{What is the total area of Europe?}

\subsection{Synthetic data generation}

To augment our training data, we generated and manually validated 1,449 additional question–answer pairs using GPT-4.1. Duplicate questions were filtered out, and common templates -- such as “how old is the person” -- were rephrased to reduce redundancy. We also followed the FreshQA style to diversify the data: the model generated both evergreen and mutable examples, with mutable questions further categorized into two subtypes. This approach enhanced the variety and coverage of our training set.

\begin{tcolorbox}[colback=gray!5!white, colframe=gray!75!black, title=Synthetic Instruction]
Can you generate different question-answer pair:
slow-changing questions, in which the answer typically changes over the course of several years (up to 10); fast-changing question, in which the answer typically changes within a year or less; never-changing, in which the answer never changes.
\end{tcolorbox}

\subsection{Short-Answer Generation Prompt} \label{appendix:short_answer_prompt}

\begin{tcolorbox}[colback=gray!5!white,
                 colframe=gray!75!black,
                 title=Short-Answer Generator Instruction]
You are a **short-answer generator**.
Given a factual **question** and a complete (possibly long) **answer**, return several *concise, semantically‑equivalent* answer variants.

\#\#\# RULES
1. Every variant must be factually correct and answer the question on its own.  

2. Keep each variant as short as possible ($\approx $1–5 words) while still unambiguous.  

3. Include the most common spellings, abbreviations, numerals $\leftrightarrow$ Roman‑numeral forms, and the canonical full form.  

4. Do **not** add information that is not explicitly in the answer.  

5. Return a JSON object **exactly** like:

\{
  "answers": [
    "Variant 1",
    "Variant 2",
    …
  ]
\}

\#\#\# EXAMPLES
Question: "Who is king of England?"
Answer: "The King of Great Britain -- Carl 3 (Charles Philip Arthur George)."
→ ["Carl 3", "King is Carl 3", "Carl III", "Charles III", "Charles Philip Arthur George"]

Question: "What is the highest mountain in the world?"
Answer: "Mount Everest is the highest mountain above sea level."
→ ["Mount Everest", "Everest", "Mt. Everest"]

Question: "Which element has the chemical symbol 'O'?"
Answer: "The chemical element with symbol O is oxygen."
→ ["oxygen", "Oxygen", "element O is oxygen"]

Question: "Who wrote the play 'Romeo and Juliet'?"
Answer: "'Romeo and Juliet' was written by William Shakespeare."
→ ["William Shakespeare", "Shakespeare"]

Question: "What is the currency of Japan?"
Answer: "The Japanese currency is the yen."
→ ["yen", "Japanese yen", "JPY"]
    
You only have to send **one message** per call.
\end{tcolorbox}

We query GPT-4o with \textsc{temperature}=0.2 and the additional tag
\texttt{"response\_format": "json\_object"} to create a short form answers from long form. It helps to better compare performance through an LLMs.

\begin{table*}[ht!]
\centering
\footnotesize
\resizebox{\textwidth}{!}{
\begin{tabular}{lllllllll}
\toprule
Model                      & Russian & English & French & German & Hebrew & Arabic & Chinese & AVG   \\
\midrule
\multicolumn{9}{c}{Validation Data (FreshQA)} \\
\midrule
BERT base cased  \\~\cite{devlin2019bertpretrainingdeepbidirectional}            & 0.822  & 0.860 & 0.800  & 0.832  & 0.770   & 0.783   & 0.854    & 0.818     \\ 
Deberta v3 base  \\~\cite{he2023debertav3improvingdebertausing}             & 0.811  & 0.851 & \textbf{0.841}  & 0.832  & \textbf{0.841}   & 0.830   & 0.834    & 0.834     \\
E5 Small \\ \cite{wang2024multilinguale5textembeddings}  & 0.809   & 0.839   & 0.818 & 0.830  & 0.801  & 0.815  & 0.794   & 0.815 \\
E5 Large \\ \cite{wang2024multilinguale5textembeddings} & \textbf{0.824}   & \textbf{0.872}   & 0.835  & \textbf{0.871}  & 0.831  & \textbf{0.835}  & \textbf{0.864}   & \textbf{0.848} \\

\midrule
\multicolumn{9}{c}{Test Data} \\
\midrule
BERT base cased  \\~\cite{devlin2019bertpretrainingdeepbidirectional}\\            & 0.893  & 0.900 & 0.889  & 0.884  & 0.889   & 0.883   & 0.902   & 0.891     \\ 
Deberta v3 base  \\~\cite{he2023debertav3improvingdebertausing}   & 0.836   & 0.842   & 0.845  & 0.841  & 0.832  & 0.825  & 0.831   & 0.836 \\
E5 Small \\ \cite{wang2024multilinguale5textembeddings}   & 0.821   & 0.822   & 0.819  & 0.815  & 0.804  & 0.807  & 0.817   & 0.815 \\
E5 Large \\ \cite{wang2024multilinguale5textembeddings}& \textbf{0.910}   & \textbf{0.913}   & \textbf{0.909}  & \textbf{0.910}  & \textbf{0.904}  & \textbf{0.900}  & \textbf{0.897}   & \textbf{0.906}    \\
\bottomrule
\end{tabular}
}
\caption{Comparison of different models on a training dataset. All models are multilingual variants. The best scores are shown in \textbf{bold}.}
\label{tab:train_e5}
\end{table*}



\section{Error Analysis Extended }

\label{appendix:error_analysis_detailed}

\begin{table*}[ht!]
\centering
\small
\setlength{\tabcolsep}{2pt}
\resizebox{\textwidth}{!}{
\begin{tabular}{ll}
\toprule
\multicolumn{1}{c}{\textbf{Misclassification reason}} & \multicolumn{1}{c}{\textbf{Example questions}} \\
\midrule

\multicolumn{2}{c}{\textit{False Positives (non-evergreen, but classified as evergreen)}} \\
\midrule

\rowcolor{violet!10}
Temporal phrasing mistaken for fixed historical facts & 
\begin{tabular}[c]{@{}l@{}} $\cdot$ In what year will the presidential election take place in Russia? \\  $\cdot$ When will the full moon be in April? \\  \end{tabular} \\

\rowcolor{violet!15}
Superlatives assumed to be static facts & 
\begin{tabular}[c]{@{}l@{}} $\cdot$ What is the biggest star in the sky? \\  $\cdot$ Which tea is the healthiest? \\  $\cdot$ What is the most popular social network in the world?\end{tabular} \\

\rowcolor{violet!20}
Biographical/life data on alive people treated as static & 
\begin{tabular}[c]{@{}l@{}}  $\cdot$ In which movies has Danila Kozlovsky acted?\\  $\cdot$ How many works has Stephen King written?\end{tabular} \\

\rowcolor{violet!25}
Geographic facts seen as immutable & 
\begin{tabular}[c]{@{}l@{}}  $\cdot$ What is the length of the Amazon River?
 \\  $\cdot$ Where is the largest zoo located?\end{tabular} \\

 \rowcolor{violet!30}
“How-to” questions with time-sensitive/legal context & 
\begin{tabular}[c]{@{}l@{}}  $\cdot$ How can maternity capital be used for building a house? 

 \\  $\cdot$ How can I contact Sberbank from abroad?
 \end{tabular} \\

\midrule
\multicolumn{2}{c}{\textit{False Negatives (evergreen, but classified as non-evergreen)}} \\
\midrule

\rowcolor{green!5}
Superlatives treated as time-sensitive or trend-based & 
\begin{tabular}[c]{@{}l@{}}  $\cdot$ What is the oldest currency? \\  $\cdot$ The rarest element in the periodic table. \\  $\cdot$ How long did the shortest war in history last?\end{tabular} \\

\rowcolor{green!12}
\begin{tabular}[c]{@{}l@{}}Biological and geographical facts wrongly assumed \\ to change frequently\end{tabular} & 
\begin{tabular}[c]{@{}l@{}}  $\cdot$ What is the area of Liechtenstein? \\  $\cdot$ Which animal has the highest blood pressure?\\  $\cdot$ How many species of elephants currently live on the planet?\end{tabular} \\

\rowcolor{green!20}
\begin{tabular}[c]{@{}l@{}}Cultural or mythological constants treated as mutable\end{tabular} & 
\begin{tabular}[c]{@{}l@{}}  $\cdot$ Where does Ded Moroz live? 
 \\  $\cdot$ How old is Ded Moroz?
 \end{tabular} \\

 \rowcolor{green!29}
\begin{tabular}[c]{@{}l@{}}Historical events treated as recent or developing stories\end{tabular} & 
\begin{tabular}[c]{@{}l@{}}  $\cdot$ In what year was the last eruption of Mount Vesuvius?
 \\  $\cdot$ What is the role of the French Revolution?
 \end{tabular} \\

 \rowcolor{green!35}
\begin{tabular}[c]{@{}l@{}}Recent years treated as too recent to be stable\end{tabular} & 
\begin{tabular}[c]{@{}l@{}}  $\cdot$ Who was recognized as the best actor in 2024?
 \\  $\cdot$ Who is in first place on the Forbes list in 2024?
\\ $\cdot$  What is the subsistence minimum set in Russia in 2024?
\\ $\cdot$ What is the most popular TV series in 2023?
 \end{tabular} \\

\bottomrule
\end{tabular}
}
\caption{Error Analysis of EG-E5 Classifier: breakdown of misclassification patterns. }
\label{tab:error_analysis_long}
\end{table*}

\section{License and Infrastructure}

All experiments were conducted using 1–2 NVIDIA A100 GPUs, totaling approximately 40 GPU-hours. Model usage adhered to their respective licenses: LLaMA 3.1~\cite{grattafiori2024llama3herdmodels} and Gemma 2~\cite{gemmateam2024gemma2improvingopen} under custom licenses, Phi 3~\cite{abdin2024phi3technicalreporthighly} and E5 under MIT, and Qwen 2.5~\cite{qwen2.5} and Mistrals~\cite{jiang2023mistral7b} under Apache 2.0. GPT Models were accessed via API or web-interface\footnote{\url{https://openai.com/}}.
We release our dataset and classifier under the MIT License.

\end{document}